\documentclass[a4paper]{article}

\usepackage{amsmath}
\usepackage{amsthm}
\usepackage{amssymb}
\usepackage{amstext}
\usepackage{mathtools} 
\usepackage{bbm}
\usepackage{graphicx}
\usepackage{subfig}
\usepackage[dvips]{epsfig}
\usepackage{texdraw}
\usepackage{comment} 
\usepackage{a4wide}
\usepackage[toc,page]{appendix} 
\usepackage{url}
\usepackage{placeins}
\usepackage{framed} 
\usepackage{hyperref} 
\usepackage{tikz}
\usetikzlibrary{shapes.misc}
\tikzset{cross/.style={cross out, draw=black, minimum size=2*(#1-\pgflinewidth), inner sep=0pt, outer sep=0pt},cross/.default={1.5pt}}

\usepackage{array,booktabs,multirow}  

\usepackage{listings}  
\usepackage{color} 
\definecolor{mygreen}{RGB}{28,172,0} 
\definecolor{mylilas}{RGB}{170,55,241}
\newcommand\norm[1]{\left\lVert#1\right\rVert}  

\let\oldref\ref
\renewcommand{\ref}[1]{(\oldref{#1})} 

\DeclareSymbolFont{msbm}{U}{msb}{m}{n}
\DeclareMathSymbol{\N}{\mathalpha}{msbm}{'116}
\DeclareMathSymbol{\R}{\mathalpha}{msbm}{'122}

\newcommand{\be}{\begin{eqnarray}}
\newcommand{\ee}{\end{eqnarray}}

\newtheorem{theorem}{Theorem}[section]
\newtheorem{lemma}[theorem]{Lemma}

\newtheorem{definition}[theorem]{Definition}

\numberwithin{equation}{section}

\begin{document}

\title{Using Low-Discrepancy Points for Data Compression in Machine Learning: An Experimental Comparison} 

\author{S. G\"ottlich\footnotemark[1], \; J. Heieck\footnotemark[1], \; A. Neuenkirch\footnotemark[1]}

\footnotetext[1]{University of Mannheim, Department of Mathematics, 68131 Mannheim, Germany. Corresponding author: Simone G\"ottlich (goettlich@math.uni-mannheim.de).}

\date{\today}

\maketitle

\begin{abstract}
Low-discrepancy points (also called Quasi-Monte Carlo points) are deterministically and cleverly chosen point sets in the unit cube, which provide an approximation of the uniform distribution.
We explore two methods based on such low-discrepancy points to reduce large data sets in order to train neural networks. The first one is the method of Dick and Feischl \cite{1}, which relies on digital nets and an averaging procedure. Motivated by our experimental findings, we construct a second method, which again uses digital nets, but Voronoi clustering instead of averaging.
Both methods are compared to the supercompress approach of \cite{supercompress}, which is a variant of  the $K$-means clustering algorithm. 
The comparison is done in terms of the compression error for different objective functions and the accuracy of the training of a neural network.
\end{abstract}



\section{Introduction}
\label{sec:1}
Data reduction is a classical technique that reduces the size of a dataset while still preserving the most important information. Concepts and methods in this field include Core-Sets \cite{CoreSets}, support points \cite{support} and random subsampling \cite{subsampling}, to mention a few. However, low-discrepancy points and quasi Monte Carlo-techniques seem to have received little attention in this context so far. They were used to create training data points for learning surrogate models in urban traffic by \cite{Cervella} and also similarly in \cite{Longo,Mishra} for a variety of applications, including several types of partial differential equations. Dick and Feischl \cite{1} proceeded differently, namely by using low-discrepancy points to compress known data for the training of neural networks. This work was the starting point of our study.

\smallskip
\smallskip

We assume that the original data, denoted by $\mathcal{X}$, is a set of $N$ points in $[0,1)^{s}$. The corresponding responses, denoted by $\mathcal{Y}$, are a set of $N$ points in $\mathbb{R}$. The objective is to predict the relationship between the $s$ attributes of the data points from $\mathcal{X}$ and the single attribute  of the responses in $\mathcal{Y}$. This is achieved through the use of a  parametrized predictor function $f_{\theta}:[0,1)^{s} \rightarrow \mathbb{R}$ where $\theta \in \Theta \subseteq \mathbb{R}^p$.
The quality of the predictor function $f_{\theta}$ is measured by the $\ell_2$-distance between the predicted value $f_{\theta}\left(\boldsymbol{x}_{n}\right)$ and the actual value $y_{n}$ for each instance $n=1, \ldots, N$, which is a classical approach for such regression problems.
Hence, the  error of the predictor $f_{\theta}$ is defined as
\begin{equation}\label{err}
	\operatorname{err}\left(f_{\theta}\right):=\frac{1}{N} \sum_{n=1}^{N}\left(f_{\theta}\left(\boldsymbol{x}_{n}\right)-y_{n}\right)^{2}.
	\stepcounter{equation}\tag{\theequation}
\end{equation}

To identify an optimal value for $\theta$, which can be chosen from the set $\Theta$, it may be necessary to evaluate the error function (or potentially its derivatives $\nabla_{\theta}^{m} \operatorname{err}\left(f_{\theta}\right)$, for values of $m\in\mathbb{N}$) on a number of occasions, resulting in a cost that is proportional to 
$$
\text { \#optimisation steps } \times \ \underbrace{\text{\#data points}}_{=N}.
$$

To reduce this cost, one can use a compressed data set, denoted by $P=\left\{\boldsymbol{z}_{1},\ldots, \boldsymbol{z}_{L}\right\}\subset[0,1)^{s}$ with $L \ll N$, in combination with an approximation of the quadratic loss function \eqref{err}.

In the case of $P$ resulting from the supercompress method from \cite{supercompress} or the QMC-Voronoi-method, each compressed data point will be assigned to a corresponding (approximate) response, denoted by $\mathcal{W} = \{w_{1},\ldots,w_{L}\}$. The approximation of the quadratic loss function is then given by 
\begin{equation}\label{eq:error_supercompress}
	\operatorname{err}(f_{\theta}) \approx \operatorname{app}_{L}^{clst}(f_{\theta}):= \frac{1}{L}\sum_{l=1}^{L} \left(f_{\theta}\left(\boldsymbol{z}_{l}\right)-w_{l}\right)^{2}.
	\stepcounter{equation}\tag{\theequation}
\end{equation}
Both approaches rely on clustering, hence we use the label {\it clst} for the approximate error function.

The approach of Dick and Feischl \cite{1} proceeds in a different way. Here, for the reduced set of responses $P$, weights $\left\{W_{\mathcal{X}, P, \nu, l}\right\}_{l=1}^{L}$ and $\left\{W_{\mathcal{X}, \mathcal{Y}, P, \nu, l}\right\}_{l=1}^{L} \subset \mathbb{R}$ are computed and  the approximation of the loss is given by
\begin{equation}\label{eq:error_QMC}
	\operatorname{err}(f_{\theta}) \approx \operatorname{app}_{L}^{avg}(f_{\theta}):=\sum_{l=1}^{L} f_{\theta}^{2}\left(\boldsymbol{z}_{l}\right) W_{\mathcal{X}, P, \nu, l}-2 \sum_{l=1}^{L} f_{\theta}\left(\boldsymbol{z}_{l}\right) W_{\mathcal{X}, \mathcal{Y}, P, \nu, l}+\frac{1}{N} \sum_{n=1}^{N} y_{n}^{2}.
	\stepcounter{equation}\tag{\theequation}
\end{equation}
Here $\nu$ is a parameter, which will be explained later on. Since the calculation of the weights relies on an averaging procedure, we use the  label {\it avg} for the approximate error function.

The reduced points sets, approximate responses and weights are all independent of  $\theta$, which implies that they can be calculated once at the outset and subsequently reused throughout the optimization process. Since  $L \ll N$, the cost is now proportional to
$$
\text { \#optimisation steps } \times \ \underbrace{\text{\#compressed data points}}_{=L}.
$$

In order to compare the approaches, we will first apply them to some given and fixed functions $f$ instead of $f_{\theta}$, see Subsection \ref{subsec:test_functions}, and will study the error of the approximate loss functions.

In a second step, we will compare the performance of the methods, when used for fitting neural networks. To this end, we will train a neural network with the original data $\mathcal{X}$ and $\mathcal{Y}$ and then compare the prediction accuracy of the original data with that of the neural network trained with the compressed set. This will be done for the MNIST data set \cite{MNIST}, as proposed in \cite{1}. Our experiments support the hypothesis that for the studied problems the adaptive clustering from the supercompress approach is superior to the QMC-Voronoi method, which again performs better than the QMC-averaging approach.

The remainder of this manuscript is structured as follows: In the next section, we will give a short motivation, why low-discrepancy point sets could be beneficial for data reduction in regression problems.
In Section \ref{sec:2} we describe the different data compression approaches, while Section \ref{sec:3} contains our numerical experiments. Our detailed findings and conclusion can be found in Section \ref{sec:4}. 

\section{Low-discrepancy point sets and regression} \label{DataCompressionMethods}

We start by introducing the concept of low-discrepancy point sets and some corresponding results. See, e.g. Chapter 2 in \cite{Niederreiter} for further information. Let $\mathcal{P}=\left\{\boldsymbol{x}_{1}, \ldots, \boldsymbol{x}_{N}\right\} \subset [0,1)^{s}
$ and 
$$
A\left( [\boldsymbol{a}, \boldsymbol{b}), \mathcal{P}\right)=\sum_{n=1}^{N} \mathbbm{1}_{[\boldsymbol{a}, \boldsymbol{b})}\left({\boldsymbol{x}}_{n}\right),
$$
where  $\mathbbm{1}_{[\boldsymbol{a},\boldsymbol{b})}$ is the characteristic function of the interval $[\boldsymbol{a},\boldsymbol{b})=\prod_{i=1}^s[a_i,b_i)$. Hence, $A\left( [\boldsymbol{a}, \boldsymbol{b}), \mathcal{P}\right)$ is the number  of the points of $\mathcal{P}$, which belong to $[\boldsymbol{a},\boldsymbol{b})$.
The discrepancy $D_{N}$ of the point set $\mathcal{P}$
is then defined as
$$
D_{N}(\mathcal{P})=\sup_{\substack{\boldsymbol{a}, \boldsymbol{b} \in [0,1]^{s} \\ \boldsymbol{a} \leq \boldsymbol{b} } }\left|\frac{A([\boldsymbol{a}, \boldsymbol{b}), \mathcal{P})}{N}-\lambda_{s}([\boldsymbol{a}, \boldsymbol{b}))\right|
$$
and measures the deviation of the empirical distribution of the points in $\mathcal{P}$ from the uniform distribution $\lambda_s$. A related quantity is the so-called star-discrepancy
$$ D_{N}^{*}(\mathcal{P})=\sup_{\boldsymbol{a}\in[0,1]^{s}}\left|\frac{A([{0}, \boldsymbol{a}), \mathcal{P})}{N}-\lambda_{s}([{0}, \boldsymbol{a}))\right|.
$$
Point sets $\mathcal{P}$ with small (star-)discrepancy are suitable for the numerical integration of functions   $g:[0,1]^s \rightarrow \mathbb{R}$. In particular, if $g$ is continuous we have the error bound
$$   \left| \frac{1}{N} \sum_{i=1}^N g(\boldsymbol{x}_i)
      - \int_{[0,1]^s} g(u)\,du \right| \leq 4 w(g; \mathcal{D}_N^* (\boldsymbol{x}_1,\ldots,\boldsymbol{x}_N)^{1/s}) $$
with the modulus of continuity
$$ w(g; \delta)= \sup_{ \substack{ \boldsymbol{u},  \boldsymbol{v} \in [0,1]^{s} \\ \| \boldsymbol{u}-\boldsymbol{v} \|_{\infty} \leq \delta}}  |g( \boldsymbol{u})-g(\boldsymbol{v})|, \quad \delta >0.$$
Moreover, the famous   Koksma-Hlawka inequality reads as
\begin{equation}\label{Koksma-Hlawka}
    \left| \frac{1}{N} \sum_{i=1}^N g(\boldsymbol{x}_i)
      - \int_{[0,1]^s} g(u)\,du \right|
     \le V(g)\, \mathcal{D}_N^* (\boldsymbol{x}_1,\ldots,\boldsymbol{x}_N).
\end{equation}
Here and in the following $V(g)$ is the Hardy-Krause variation of $g$. 

Now, let $X$ be a random vector with values in $[0,1]^s$, $Y$ be  a random variable and $f_{\theta}:[0,1]^s\rightarrow \mathbb{R}$ with  $\theta \in \Theta \subseteq \mathbb{R}^p$. Finding the parameter $\theta$, which minimizes the 
expected prediction error
$$ \mathcal{L}(\theta)=\mathbf{E}|f_{\theta}(X)-Y|^2,$$
is the classical $L^2$-regression problem.
The loss function \eqref{err}, i.e.
\begin{equation*}
	\operatorname{err}\left(f_{\theta}\right):=\frac{1}{N} \sum_{n=1}^{N}\left(f_{\theta}\left(\boldsymbol{x}_{n}\right)-y_{n}\right)^{2},
\end{equation*}
can be seen as the empirical variant of the expected prediction error by assuming that
$(\boldsymbol{x}_n,y_n)$, $n=1, \ldots, N$, are independent and identically distributed realizations of $(X,Y)$. 
If $(X,Y)$ has a joint Lebesgue-density $\varphi:[0,1]^s \times \mathbb{R} \rightarrow [0, \infty)$ with marginal densities $\varphi_X: [0,1]^s \rightarrow [0, \infty)$ and $\varphi_Y:\mathbb{R} \rightarrow [0, \infty)$, then we have
\begin{align*} \mathcal{L}(\theta) & = \int_{[0,1]^s\times \mathbb{R}}    |f_{\theta}(x)-y|^2 \varphi(x,y) d(x,y) \\ & =\int_{[0,1]^s}    f_{\theta}(x)^2 \varphi_X(x) dx -2 \int_{[0,1]^s\times \mathbb{R}}  f_{\theta}(x)y \varphi(x,y) d(x,y)  + \int_{\mathbb{R}}  y^2 \varphi_Y(y) dy 
.    \end{align*}
Now using low-discrepancy points as the compressed data points $P=\left\{\boldsymbol{z}_{1},\ldots, \boldsymbol{z}_{L}\right\}\subset[0,1)^{s}$ for the $dx$-integration and the data from $\mathcal{Y}$ for the $dy$-integration, one obtains an approximation of the form \eqref{eq:error_QMC}, i.e.
\begin{equation*}
\mathcal{L}(\theta)
	 \approx \operatorname{app}_{L}^{avg}(f_{\theta})=\sum_{l=1}^{L} f_{\theta}^{2}\left(\boldsymbol{z}_{l}\right) W_{\mathcal{X}, P, \nu, l}-2 \sum_{l=1}^{L} f_{\theta}\left(\boldsymbol{z}_{l}\right) W_{\mathcal{X}, \mathcal{Y}, P, \nu, l}+\frac{1}{N} \sum_{n=1}^{N} y_{n}^{2}.
\end{equation*}
Thus, if the empirical loss function \eqref{err} is close to the expected prediction error $\mathcal{L}(\theta)$, which is reasonable if $N$ is large, then $\operatorname{app}_{L}^{avg}(f_{\theta})$ should provide a good approximation of \eqref{err}.

\section{Data compression methods}\label{DataCompressionMethods}
\label{sec:2}

In this section we present the different compression algorithms and provide the principal ideas of their implementation.

\subsection{Quasi-Monte Carlo compression}\label{sec:weighted_digital_nets}

Here we present the algorithm and results of \cite{1}, where digital nets, which are particular $(t_{\alpha},m,s)$-nets, are used as low-discrepancy sets. In the following, 
we assume that $\alpha\geq1$ is an integer and that $b \geq 2$ is prime. For a vector $\boldsymbol{d}$, $\left| \boldsymbol{d} \right|$ denotes its $\ell_{1}$-norm, while for a set $A$, the notation $\left| A \right|$ denotes its  cardinality. 

\begin{definition}[e.g., p.5,\ \cite{1}]\label{tms-net}
	For $\boldsymbol{d}\in \mathbb{N}_{0}^{s}$ we set $K_{\boldsymbol{d}} := \left\{\boldsymbol{a}=\left(a_{1}, \ldots, a_{s}\right)^{\intercal} \in \mathbb{N}_{0}^{s}: a_{j}<b^{d_{j}}\right\}.$
	A point set $P=\left\{\boldsymbol{z}_{1},\ldots, \boldsymbol{z}_{L}\right\}\subset[0,1)^{s}$ consisting of $L=b^{m}$ points is called a $(t_{\alpha},m,s)$-net in base $b$ of order $\alpha$, if every elementary interval $I_{\boldsymbol{a}, \boldsymbol{d}}=\prod_{j=1}^{s}\left[\frac{a_{j}}{b^{d_{j}}}, \frac{a_{j}+1}{b^{d_{j}}}\right)$ with $\left| \boldsymbol{d} \right|=m-t_{\alpha}$ and $\boldsymbol{a}\in K_{ \boldsymbol{d}}$ contains exactly $b^{t_{\alpha}}$ points. 
\end{definition}

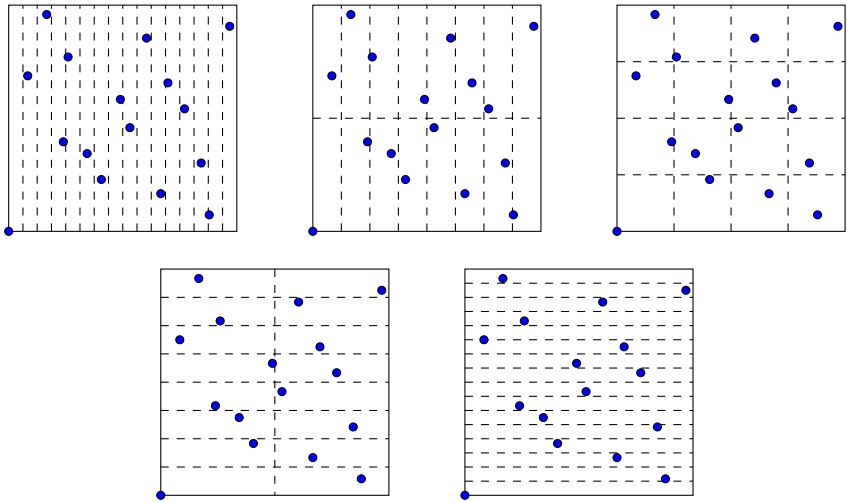
\begin{figure}[h]
	\centering
	\begin{tikzpicture}
		\draw (0,0)--(0,3)--(3,3)--(3,0)--(0,0);
		\draw[style=dashed] (3/16,0)--(3/16,3);
		\draw[style=dashed] (6/16,0)--(6/16,3);
		\draw[style=dashed] (9/16,0)--(9/16,3);
		\draw[style=dashed] (12/16,0)--(12/16,3);
		\draw[style=dashed] (15/16,0)--(15/16,3);
		\draw[style=dashed] (18/16,0)--(18/16,3);
		\draw[style=dashed] (21/16,0)--(21/16,3);
		\draw[style=dashed] (24/16,0)--(24/16,3);
		\draw[style=dashed] (27/16,0)--(27/16,3);
		\draw[style=dashed] (30/16,0)--(30/16,3);
		\draw[style=dashed] (33/16,0)--(33/16,3);
		\draw[style=dashed] (36/16,0)--(36/16,3);
		\draw[style=dashed] (39/16,0)--(39/16,3);
		\draw[style=dashed] (42/16,0)--(42/16,3);
		\draw[style=dashed] (45/16,0)--(45/16,3);
		\draw[fill=blue] (0,0) circle (1.5pt);
		\draw[fill=blue] (4/16,33/16) circle (1.5pt);
		\draw[fill=blue] (8/16,46/16) circle (1.5pt);
		\draw[fill=blue] (11.5/16,19/16) circle (1.5pt);
		\draw[fill=blue] (12.5/16,37/16) circle (1.5pt);
		\draw[fill=blue] (16.5/16,16.5/16) circle (1.5pt);
		\draw[fill=blue] (19.5/16,11/16) circle (1.5pt);
		\draw[fill=blue] (23.5/16,28/16) circle (1.5pt);
		\draw[fill=blue] (25.5/16,22/16) circle (1.5pt);
		\draw[fill=blue] (29/16,41/16) circle (1.5pt);
		\draw[fill=blue] (32/16,8/16) circle (1.5pt);
		\draw[fill=blue] (33.5/16,31.5/16) circle (1.5pt);
		\draw[fill=blue] (37/16,26/16) circle (1.5pt);
		\draw[fill=blue] (40.5/16,14.5/16) circle (1.5pt);
		\draw[fill=blue] (42.2/16,3.5/16) circle (1.5pt);
		\draw[fill=blue] (46.5/16,43.5/16) circle (1.5pt);
		
		\draw (4,0)--(4,3)--(7,3)--(7,0)--(4,0);	
		\draw[style=dashed] (6/16+4,0)--(6/16+4,3);
		\draw[style=dashed] (12/16+4,0)--(12/16+4,3);
		\draw[style=dashed] (18/16+4,0)--(18/16+4,3);
		\draw[style=dashed] (24/16+4,0)--(24/16+4,3);
		\draw[style=dashed] (30/16+4,0)--(30/16+4,3);
		\draw[style=dashed] (36/16+4,0)--(36/16+4,3);
		\draw[style=dashed] (42/16+4,0)--(42/16+4,3);
		\draw[style=dashed] (4,1.5)--(7,1.5);
		\draw[fill=blue] (4,0) circle (1.5pt);
		\draw[fill=blue] (4/16+4,33/16) circle (1.5pt);
		\draw[fill=blue] (8/16+4,46/16) circle (1.5pt);
		\draw[fill=blue] (11.5/16+4,19/16) circle (1.5pt);
		\draw[fill=blue] (12.5/16+4,37/16) circle (1.5pt);
		\draw[fill=blue] (16.5/16+4,16.5/16) circle (1.5pt);
		\draw[fill=blue] (19.5/16+4,11/16) circle (1.5pt);
		\draw[fill=blue] (23.5/16+4,28/16) circle (1.5pt);
		\draw[fill=blue] (25.5/16+4,22/16) circle (1.5pt);
		\draw[fill=blue] (29/16+4,41/16) circle (1.5pt);
		\draw[fill=blue] (32/16+4,8/16) circle (1.5pt);
		\draw[fill=blue] (33.5/16+4,31.5/16) circle (1.5pt);
		\draw[fill=blue] (37/16+4,26/16) circle (1.5pt);
		\draw[fill=blue] (40.5/16+4,14.5/16) circle (1.5pt);
		\draw[fill=blue] (42.2/16+4,3.5/16) circle (1.5pt);
		\draw[fill=blue] (46.5/16+4,43.5/16) circle (1.5pt);
		
		\draw (8,0)--(8,3)--(11,3)--(11,0)--(8,0);
		\draw[style=dashed] (8.75,0)--(8.75,3);
		\draw[style=dashed] (9.5,0)--(9.5,3);
		\draw[style=dashed] (10.25,0)--(10.25,3);
		\draw[style=dashed] (8,1.5)--(11,1.5);	
		\draw[style=dashed] (8,0.75)--(11,0.75);	
		\draw[style=dashed] (8,2.25)--(11,2.25);	
		\draw[fill=blue] (8,0) circle (1.5pt);
		\draw[fill=blue] (4/16+8,33/16) circle (1.5pt);
		\draw[fill=blue] (8/16+8,46/16) circle (1.5pt);
		\draw[fill=blue] (11.5/16+8,19/16) circle (1.5pt);
		\draw[fill=blue] (12.5/16+8,37/16) circle (1.5pt);
		\draw[fill=blue] (16.5/16+8,16.5/16) circle (1.5pt);
		\draw[fill=blue] (19.5/16+8,11/16) circle (1.5pt);
		\draw[fill=blue] (23.5/16+8,28/16) circle (1.5pt);
		\draw[fill=blue] (25.5/16+8,22/16) circle (1.5pt);
		\draw[fill=blue] (29/16+8,41/16) circle (1.5pt);
		\draw[fill=blue] (32/16+8,8/16) circle (1.5pt);
		\draw[fill=blue] (33.5/16+8,31.5/16) circle (1.5pt);
		\draw[fill=blue] (37/16+8,26/16) circle (1.5pt);
		\draw[fill=blue] (40.5/16+8,14.5/16) circle (1.5pt);
		\draw[fill=blue] (42.2/16+8,3.5/16) circle (1.5pt);
		\draw[fill=blue] (46.5/16+8,43.5/16) circle (1.5pt);
		
		\draw (2,-3.5)--(5,-3.5)--(5,-0.5)--(2,-0.5)--(2,-3.5);
		\draw[style=dashed] (2,-2)--(5,-2);
		\draw[style=dashed] (2,-1.25)--(5,-1.25);
		\draw[style=dashed] (2,-2.75)--(5,-2.75);
		\draw[style=dashed] (2,-2.75/2-3.5/2)--(5,-2.75/2-3.5/2);
		\draw[style=dashed] (2,-2.75/2-2/2)--(5,-2.75/2-2/2);
		\draw[style=dashed] (2,-1.25/2-2/2)--(5,-1.25/2-2/2);
		\draw[style=dashed] (2,-1.25/2-0.5/2)--(5,-1.25/2-0.5/2);
		\draw[style=dashed] (3.5,-0.5)--(3.5,-3.5);
		\draw[fill=blue] (2,-3.5) circle (1.5pt);
		\draw[fill=blue] (4/16+2,33/16-3.5) circle (1.5pt);
		\draw[fill=blue] (8/16+2,46/16-3.5) circle (1.5pt);
		\draw[fill=blue] (11.5/16+2,19/16-3.5) circle (1.5pt);
		\draw[fill=blue] (12.5/16+2,37/16-3.5) circle (1.5pt);
		\draw[fill=blue] (16.5/16+2,16.5/16-3.5) circle (1.5pt);
		\draw[fill=blue] (19.5/16+2,11/16-3.5) circle (1.5pt);
		\draw[fill=blue] (23.5/16+2,28/16-3.5) circle (1.5pt);
		\draw[fill=blue] (25.5/16+2,22/16-3.5) circle (1.5pt);
		\draw[fill=blue] (29/16+2,41/16-3.5) circle (1.5pt);
		\draw[fill=blue] (32/16+2,8/16-3.5) circle (1.5pt);
		\draw[fill=blue] (33.5/16+2,31.5/16-3.5) circle (1.5pt);
		\draw[fill=blue] (37/16+2,26/16-3.5) circle (1.5pt);
		\draw[fill=blue] (40.5/16+2,14.5/16-3.5) circle (1.5pt);
		\draw[fill=blue] (42.2/16+2,3.5/16-3.5) circle (1.5pt);
		\draw[fill=blue] (46.5/16+2,43.5/16-3.5) circle (1.5pt);
		
		\draw (6,-3.5)--(9,-3.5)--(9,-0.5)--(6,-0.5)--(6,-3.5);	
		\draw[style=dashed] (6,-2)--(9,-2);
		\draw[style=dashed] (6,-1.25)--(9,-1.25);
		\draw[style=dashed] (6,-2.75)--(9,-2.75);
		\draw[style=dashed] (6,-2.75/2-3.5/2)--(9,-2.75/2-3.5/2);
		\draw[style=dashed] (6,-2.75/2-2/2)--(9,-2.75/2-2/2);
		\draw[style=dashed] (6,-1.25/2-2/2)--(9,-1.25/2-2/2);
		\draw[style=dashed] (6,-1.25/2-0.5/2)--(9,-1.25/2-0.5/2);	
		\draw[style=dashed] (6,-0.5-3/16)--(9,-0.5-3/16);
		\draw[style=dashed] (6,-0.5-9/16)--(9,-0.5-9/16);
		\draw[style=dashed] (6,-0.5-15/16)--(9,-0.5-15/16);
		\draw[style=dashed] (6,-0.5-21/16)--(9,-0.5-21/16);
		\draw[style=dashed] (6,-0.5-27/16)--(9,-0.5-27/16);
		\draw[style=dashed] (6,-0.5-33/16)--(9,-0.5-33/16);
		\draw[style=dashed] (6,-0.5-39/16)--(9,-0.5-39/16);
		\draw[style=dashed] (6,-0.5-45/16)--(9,-0.5-45/16);
		\draw[fill=blue] (6,-3.5) circle (1.5pt);
		\draw[fill=blue] (4/16+6,33/16-3.5) circle (1.5pt);
		\draw[fill=blue] (8/16+6,46/16-3.5) circle (1.5pt);
		\draw[fill=blue] (11.5/16+6,19/16-3.5) circle (1.5pt);
		\draw[fill=blue] (12.5/16+6,37/16-3.5) circle (1.5pt);
		\draw[fill=blue] (16.5/16+6,16.5/16-3.5) circle (1.5pt);
		\draw[fill=blue] (19.5/16+6,11/16-3.5) circle (1.5pt);
		\draw[fill=blue] (23.5/16+6,28/16-3.5) circle (1.5pt);
		\draw[fill=blue] (25.5/16+6,22/16-3.5) circle (1.5pt);
		\draw[fill=blue] (29/16+6,41/16-3.5) circle (1.5pt);
		\draw[fill=blue] (32/16+6,8/16-3.5) circle (1.5pt);
		\draw[fill=blue] (33.5/16+6,31.5/16-3.5) circle (1.5pt);
		\draw[fill=blue] (37/16+6,26/16-3.5) circle (1.5pt);
		\draw[fill=blue] (40.5/16+6,14.5/16-3.5) circle (1.5pt);
		\draw[fill=blue] (42.2/16+6,3.5/16-3.5) circle (1.5pt);
		\draw[fill=blue] (46.5/16+6,43.5/16-3.5) circle (1.5pt);
	\end{tikzpicture}
	\caption{$(0, 4, 2)$-net in base $2$ (blue points)}
	\label{042-Netz}
\end{figure}

The idea behind such nets is to identify a set of points that is evenly distributed within the unit cube $[0,1)^{s}$. It is essential that each elementary interval of size $b^{-(m-t_{\alpha})}$ contains an identical number of points. Figure \ref{042-Netz} shows an example of a first order $(0,4,2)$-net in base $b=2$. The point set is well distributed for every partition of the unit square into elementary intervals of size $b^{-4}=\frac{1}{16}$. This means that every interval contains the same number of points.

For each $(t_{\alpha},m,s)$-net $P$ in base $b$ of order $\alpha$ and for each $\nu\leq m-t_{\alpha}$, $\boldsymbol{d} \in \mathbb{N}_{0}^{s}$ and $\boldsymbol{a}\in K_{ \boldsymbol{d}}$, it can be shown that $\left| P_{\boldsymbol{a}, \boldsymbol{d}} \right|=b^{m-\nu+q}$, where $P_{\boldsymbol{a}, \boldsymbol{d}}$ denotes the set $P$ intersected with $I_{\boldsymbol{a},\boldsymbol{d}} $ and $\left| \boldsymbol{d}\right|=\nu-q$ for $q \in \{0,\ldots,\min(s-1,\nu)\}$. As mentioned in Chapter 2 of \cite{1}, this is a consequence of Definition \ref{tms-net} and will be important for deriving the weights.

\smallskip
\smallskip

We  need to introduce further notations. For $\boldsymbol{y} \in[0,1)^{s}$, $I_{\boldsymbol{d}}(\boldsymbol{y})$ describes the elementary interval $I_{\boldsymbol{a}, \boldsymbol{d}}$, which contains $\boldsymbol{y}$. We define the set of points that lie within the given elementary interval as follows: 
$$\mathcal{X}_{\boldsymbol{a}, \boldsymbol{d}} := \mathcal{X} \cap I_{\boldsymbol{a}, \boldsymbol{d}}, \quad  \mathcal{X}_{\boldsymbol{d}}(\boldsymbol{y}):=\mathcal{X} \cap I_{\boldsymbol{d}}(\boldsymbol{y}) \quad \text{and} \quad P_{\boldsymbol{d}}(\boldsymbol{y}):=P \cap I_{\boldsymbol{d}}(\boldsymbol{y}).$$
Furthermore, we define the set $K_{\nu}$ as the union of all $K_{\boldsymbol{d}}$ for which $\boldsymbol{d} \in \mathbb{N}_{0}^{s}$ and $\lvert \boldsymbol{d} \rvert=\nu$. The volume of each elementary interval is determined by the value of $\nu$, which is used to calculate the weights. 

With these notations in hand, we state a Lemma, which will be used to calculate the weights by an averaging procedure, and allows us to work with $K_{\nu}$. This is done in order to include every partition of the unit cube, for which the elementary intervals exhibit the same volume.

\begin{lemma}[Lemma 1,\ \cite{1}]\label{komprin}
	Let $\nu \geq 0$ be an integer. For all $\boldsymbol{a} \in \mathbb{N}_{0}^{s}$ the combination principle 
	$$
	\mathbbm{1}_{K_{\nu}}(\boldsymbol{a}) = \sum_{q=0}^{\min(s-1,\nu)}(-1)^{q}\binom{s-1}{q} \sum_{\substack{\boldsymbol{d} \in \mathbb{N}_{0}^{s} \\ \lvert \boldsymbol{d} \rvert =\nu-q}} \mathbbm{1}_{K_{\boldsymbol{d}}}(\boldsymbol{a})
	$$
    holds. Here, $\mathbbm{1}_{A}$ denotes the indicator function for an arbitrary set $A$.
\end{lemma}

Lemma \ref{komprin} can be exploited to find an expression for the weights $\left\{W_{\mathcal{X}, P, \nu, l}\right\}_{l=1}^{L}$. In the following, we assume that our data set is given by $\mathcal{X}=\left\{\boldsymbol{x}_{1}, \ldots, \boldsymbol{x}_{N}\right\} $, our responses by $\mathcal{Y}=\left\{y_{1}, \ldots, y_{N}\right\}$, our compressed data set by $P=\left\{\boldsymbol{z}_{1},\ldots, \boldsymbol{z}_{L}\right\}$, and that $\nu$ is fixed.

Firstly, we consider a fixed $\boldsymbol{d} \in \mathbb{N}_{0}^{s}$, which represents a fixed partition of the unit cube. This allows us to derive the following approximation:
\begin{equation*}
	\frac{1}{N} \sum_{n=1}^{N} f_{\theta}^{2}\left(\boldsymbol{x}_{n}\right) \overset{}{\approx}\sum_{l=1}^{L} f_{\theta}^{2}\left(\boldsymbol{z}_{l}\right) \frac{\left|\mathcal{X}_{\boldsymbol{d}}\left(\boldsymbol{z}_{l}\right)\right|}{N} \frac{1}{\left|P_{\boldsymbol{d}}\left(\boldsymbol{z}_{l}\right)\right|} .
\end{equation*}
If the function $f_{\theta}$ is constant on each of the elementary intervals provided by the fixed partition, equality is achieved, since $\left|\mathcal{X}_{\boldsymbol{d}}\left(\boldsymbol{z}_{l}\right)\right|$ is the number of data points from $\mathcal{X}$, which are in $I_d(\boldsymbol{z}_{l})$, and  $\left|P_{\boldsymbol{d}}\left(\boldsymbol{z}_{l}\right)\right|$ is the number of  compressed data points from $P$, which are in $I_d(\boldsymbol{z}_{l})$. Consequently, dividing by this number removes multiple counting. For regular functions, the approximation should perform well.
Conversely, the greater the variation in function values on each interval, the worse the approximation will be.

Rather than focusing on a single partition, \cite{1} considers all partitions with volume $b^{-\nu}$. Therefore, we must average over all partitions $\boldsymbol{d} \in \mathbb{N}_{0}^{s}$ with $\left| \boldsymbol{d} \right| = \nu$ and apply Lemma \ref{komprin}:
\begin{equation*}
	\frac{1}{N} \sum_{n=1}^{N} f_{\theta}^{2}\left(\boldsymbol{x}_{n}\right) \overset{}{\approx} \sum_{l=1}^{L} f_{\theta}^{2}\left(\boldsymbol{z}_{l}\right) \underbrace{\sum_{q=0}^{\min(s-1,\nu)}(-1)^{q} \binom{s-1}{q} \sum_{\substack{\boldsymbol{d} \in \mathbb{N}_{0}^{s} \\
			|\boldsymbol{d}|=\nu-q}} \frac{\left|\mathcal{X}_{\boldsymbol{d}}\left(\boldsymbol{z}_{l}\right)\right|}{N} \frac{1}{\left|P_{\boldsymbol{d}}\left(\boldsymbol{z}_{l}\right)\right|}}_{=W_{\mathcal{X}, P, \nu, l}}.
\end{equation*}
The quality of this approximation depends strongly on $P$ and $\nu$, and its efficacy will be evaluated at a later stage. For the time being, we merely motivate the selection of weights.
In the above equation the ratio $\frac{\left|\mathcal{X}_{\boldsymbol{d}}\left(\boldsymbol{z}_{l}\right)\right|} {\left|P_{\boldsymbol{d}}\left(\boldsymbol{z}_{l}\right)\right|}$ reflects the relevance of $\boldsymbol{z}_{l}$. The greater the number of points in the set $\mathcal{X}$ that fall within the same elementary interval as the point $\boldsymbol{z}_{l}$ for a multitude of partitions, the higher the weight assigned to $\boldsymbol{z}_{l}$. Consequently, the quantity $f_{\theta}^{2}\left(\boldsymbol{z}_{l}\right)$ will be of greater significance in the approximation.  

Assuming that  $P$ is a $(t_{\alpha},m,s)$-net in base $b$ of order $\alpha$, we can exploit the $(t_{\alpha},m,s)$-net property, namely the fact that $\left| P_{\boldsymbol{a}, \boldsymbol{d}} \right|=b^{m-\nu+q}$, where $\left| \boldsymbol{d}\right|=\nu-q$ for $q \in \{0,\ldots,\min(s-1,\nu)\}$. This simplifies the weights to the following expression given in \cite{1}:
\begin{equation}\label{weights_1}
W_{\mathcal{X}, P, \nu, l}=\frac{b^{\nu-m}}{N} \sum_{q=0}^{\min(s-1,\nu)}(-1)^{q}\binom{s-1}{q} \frac{1}{b^{q}} \sum_{\substack{\boldsymbol{d} \in \mathbb{N}_{0}^{s} \\ |\boldsymbol{d}|=\nu-q}}\left|\mathcal{X}_{\boldsymbol{d}}\left(\boldsymbol{z}_{l}\right)\right|.
\end{equation}
Note that in order to use this representation of the weights, it is necessary to ensure that $m-t_{\alpha} \geq \nu$.

\smallskip
\smallskip

The derivation of the weights $\left\{W_{\mathcal{X}, \mathcal{Y}, P, \nu, l}\right\}_{l=1}^{L}$ is analogous. Once again, we consider a fixed partition $\boldsymbol{d} \in \mathbb{N}_{0}^{s}$ first and obtain an approximation:
\begin{equation*} 
	\frac{1}{N} \sum_{n=1}^{N} y_{n} f_{\theta}\left(\boldsymbol{x}_{n}\right) \overset{}{\approx} \sum_{l=1}^{L} f_{\theta}\left(\boldsymbol{z}_{l}\right) \frac{1}{N} \frac{1}{\left|P_{\boldsymbol{d}}\left(\boldsymbol{z}_{l}\right)\right|} \sum_{\substack{n=1 \\ \boldsymbol{x}_{n} \in I_{\boldsymbol{d}}\left(\boldsymbol{z}_{l}\right)}}^{N} y_{n}.
\end{equation*}
Considering all possible partitions with a volume of $b^{-\nu}$, the following result is obtained:
\begin{equation*} 
\frac{1}{N} \sum_{n=1}^{N} y_{n}f_{\theta}\left(\boldsymbol{x}_{n}\right) 
\approx \sum_{l=1}^{L} f_{\theta}\left(\boldsymbol{z}_{l}\right) \underbrace{\sum_{q=0}^{\min(s-1,\nu)}(-1)^{q}\binom{s-1}{q}\sum_{\substack{\boldsymbol{d} \in \mathbb{N}_{0}^{s} \\
			|\boldsymbol{d}|=\nu-q}} \frac{1}{N} \frac{1}{\left|P_{\boldsymbol{d}}\left(\boldsymbol{z}_{l}\right)\right|} \sum_{\substack{n=1 \\ \boldsymbol{x}_{n} \in I_{\boldsymbol{d}}\left(\boldsymbol{z}_{l}\right)}}^{N} y_{n}}_{=W_{\mathcal{X}, \mathcal{Y}, P, \nu, l}}.
\end{equation*}
Once more, the quantity $\frac{1}{\left|P_{\boldsymbol{d}}\left(\boldsymbol{z}_{l}\right)\right|} \sum_{\substack{n=1, \boldsymbol{x}_{n} \in I_{\boldsymbol{d}}\left(\boldsymbol{z}_{l}\right)}}^{N} y_{n}$ represents the relevance of $\boldsymbol{z}_{l}$ in  the same manner as previously. Concurrently, it shall approximate $y_{n}$. The greater the number of points in the same elementary interval as $\boldsymbol{z}_{l}$ and the larger the response values $y_{n}$, the greater the influence of $f_{\theta}\left(\boldsymbol{z}_{l}\right)$ in the approximation.

Again, if  $P$ is a $(t_{\alpha},m,s)$-net in base $b$ of order $\alpha$ and if $m-t_{\alpha} \geq \nu$, then the weights simplify to
\begin{equation}\label{weights_2}
W_{\mathcal{X}, \mathcal{Y}, P, \nu, l}=\frac{b^{\nu-m}}{N} \sum_{q=0}^{\min(s-1,\nu)}(-1)^{q}\binom{s-1}{q} \frac{1}{b^{q}} \sum_{\substack{\boldsymbol{d} \in \mathbb{N}_{0}^{s} \\ |\boldsymbol{d}|=\nu-q}} \sum_{\substack{n=1 \\
\boldsymbol{x}_{n} \in I_{\boldsymbol{d}}\left(\boldsymbol{z}_{l}\right)}}^{N} y_{n}.
\end{equation}

\smallskip
\smallskip

\begin{figure}[h]
	\centering
	\begin{tikzpicture}
		\draw (0,0)--(0,3)--(3,3)--(3,0)--(0,0);
		\draw[style=dashed] (3/16,0)--(3/16,3);
		\draw[style=dashed] (6/16,0)--(6/16,3);
		\draw[style=dashed] (9/16,0)--(9/16,3);
		\draw[style=dashed] (12/16,0)--(12/16,3);
		\draw[style=dashed] (15/16,0)--(15/16,3);
		\draw[style=dashed] (18/16,0)--(18/16,3);
		\draw[style=dashed] (21/16,0)--(21/16,3);
		\draw[style=dashed] (24/16,0)--(24/16,3);
		\draw[style=dashed] (27/16,0)--(27/16,3);
		\draw[style=dashed] (30/16,0)--(30/16,3);
		\draw[style=dashed] (33/16,0)--(33/16,3);
		\draw[style=dashed] (36/16,0)--(36/16,3);
		\draw[style=dashed] (39/16,0)--(39/16,3);
		\draw[style=dashed] (42/16,0)--(42/16,3);
		\draw[style=dashed] (45/16,0)--(45/16,3);
		\draw[fill=blue] (0,0) circle (1.5pt);
		\draw[fill=blue] (4/16,33/16) circle (1.5pt);
		\draw[fill=blue] (8/16,46/16) circle (1.5pt);
		\draw[fill=blue] (11.5/16,19/16) circle (1.5pt);
		\draw[fill=blue] (12.5/16,37/16) circle (1.5pt);
		\draw[fill=blue] (16.5/16,16.5/16) circle (1.5pt);
		\draw[fill=blue] (19.5/16,11/16) circle (1.5pt);
		\draw[fill=blue] (23.5/16,28/16) circle (1.5pt);
		\draw[fill=blue] (25.5/16,22/16) circle (1.5pt);
		\draw[fill=blue] (29/16,41/16) circle (1.5pt);
		\draw[fill=blue] (32/16,8/16) circle (1.5pt);
		\draw[fill=blue] (33.5/16,31.5/16) circle (1.5pt);
		\draw[fill=blue] (37/16,26/16) circle (1.5pt);
		\draw[fill=blue] (40.5/16,14.5/16) circle (1.5pt);
		\draw[fill=blue] (42.2/16,3.5/16) circle (1.5pt);
		\draw[fill=blue] (46.5/16,43.5/16) circle (1.5pt);
		\draw (0/16,8/16) node[cross,red] {};
		\draw (10/16,18/16) node[cross,red] {};
		\draw (43/16,22/16) node[cross,red] {};
		\draw (19/16,7/16) node[cross,red] {};
		\draw (1/16,43/16) node[cross,red] {};
		\draw (30/16,30/16) node[cross,red] {};
		\draw (22/16,2/16) node[cross,red] {};
		\draw (45/16,47.5/16) node[cross,red] {};
		\draw (8/16,26/16) node[cross,red] {};
		\draw (7/16,7/16) node[cross,red] {};
		\draw (24/16,16/16) node[cross,red] {};
		\draw (32/16,19/16) node[cross,red] {};
		\draw (39/16,28/16) node[cross,red] {};
		\draw (2/16,45/16) node[cross,red] {};
		\draw (18/16,2/16) node[cross,red] {};
		\draw (42/16,8/16) node[cross,red] {};
		\draw (33/16,1/16) node[cross,red] {};
		\draw (24/16,40/16) node[cross,red] {};
		\draw (19/16,46/16) node[cross,red] {};
		\draw (30/16,36/16) node[cross,red] {};
		\draw (2/16,22/16) node[cross,red] {};
		\draw (45/16,11/16) node[cross,red] {};
		\draw (15/16,40/16) node[cross,red] {};
		\draw (29/16,39/16) node[cross,red] {};
		\draw (11/16,45/16) node[cross,red] {};
		\draw (34/16,7/16) node[cross,red] {};
		\draw (40/16,40/16) node[cross,red] {};
		\draw (39/16,44/16) node[cross,red] {};
		\draw (38/16,39/16) node[cross,red] {};
		\draw (16/16,29/16) node[cross,red] {};
		\draw (12/16,33/16) node[cross,red] {};
		\draw (18/16,24/16) node[cross,red] {};
		\draw (15/16,15/16) node[cross,red] {};
		\draw (22/16,36/16) node[cross,red] {};
		\draw (24/16,40/16) node[cross,red] {};
		\draw (5/16,30/16) node[cross,red] {};
		\draw (8/16,18/16) node[cross,red] {};
		\draw (7/16,32/16) node[cross,red] {};
		\draw (30/16,10/16) node[cross,red] {};
		\draw (36/16,43/16) node[cross,red] {};
		\draw (16/16,8/16) node[cross,red] {};
		\draw (6/16,10/16) node[cross,red] {};
		\draw (26/16,12/16) node[cross,red] {};
		\draw (45/16,23.5/16) node[cross,red] {};
		\draw (46/16,20.5/16) node[cross,red] {};
		\draw (47/16,19.5/16) node[cross,red] {};
		\draw (47.4/16,16.5/16) node[cross,red] {};
		\draw (47.2/16,12.5/16) node[cross,red] {};
		\draw (46.4/16,8.5/16) node[cross,red] {};
		\draw (45.3/16,1.5/16) node[cross,red] {};
		\draw (45.9/16,0.5/16) node[cross,red] {};
		\draw (46.2/16,8.5/16) node[cross,red] {};
		\draw (45/16,22/16) node[cross,red] {};
		\draw (45.8/16,6/16) node[cross,red] {};
		\draw (36/16,42.1/16) node[cross,red] {};
		\draw (35.5/16,44.9/16) node[cross,red] {};
		\draw (32/16,42.5/16) node[cross,red] {};
		\draw (31/16,42.9/16) node[cross,red] {};
		\draw (31.5/16,43/16) node[cross,red] {};
		\draw (34.4/16,43.9/16) node[cross,red] {};
		\draw (34.6/16,44.5/16) node[cross,red] {};
		\draw (29/16,44.7/16) node[cross,red] {};
		
		\draw (4,0)--(4,3)--(7,3)--(7,0)--(4,0);	
		\draw[style=dashed] (6/16+4,0)--(6/16+4,3);
		\draw[style=dashed] (12/16+4,0)--(12/16+4,3);
		\draw[style=dashed] (18/16+4,0)--(18/16+4,3);
		\draw[style=dashed] (24/16+4,0)--(24/16+4,3);
		\draw[style=dashed] (30/16+4,0)--(30/16+4,3);
		\draw[style=dashed] (36/16+4,0)--(36/16+4,3);
		\draw[style=dashed] (42/16+4,0)--(42/16+4,3);
		\draw[style=dashed] (4,1.5)--(7,1.5);
		\draw[fill=blue] (4,0) circle (1.5pt);
		\draw[fill=blue] (4/16+4,33/16) circle (1.5pt);
		\draw[fill=blue] (8/16+4,46/16) circle (1.5pt);
		\draw[fill=blue] (11.5/16+4,19/16) circle (1.5pt);
		\draw[fill=blue] (12.5/16+4,37/16) circle (1.5pt);
		\draw[fill=blue] (16.5/16+4,16.5/16) circle (1.5pt);
		\draw[fill=blue] (19.5/16+4,11/16) circle (1.5pt);
		\draw[fill=blue] (23.5/16+4,28/16) circle (1.5pt);
		\draw[fill=blue] (25.5/16+4,22/16) circle (1.5pt);
		\draw[fill=blue] (29/16+4,41/16) circle (1.5pt);
		\draw[fill=blue] (32/16+4,8/16) circle (1.5pt);
		\draw[fill=blue] (33.5/16+4,31.5/16) circle (1.5pt);
		\draw[fill=blue] (37/16+4,26/16) circle (1.5pt);
		\draw[fill=blue] (40.5/16+4,14.5/16) circle (1.5pt);
		\draw[fill=blue] (42.2/16+4,3.5/16) circle (1.5pt);
		\draw[fill=blue] (46.5/16+4,43.5/16) circle (1.5pt);
		\draw (0/16+4,8/16) node[cross,red] {};
		\draw (10/16+4,18/16) node[cross,red] {};
		\draw (43/16+4,22/16) node[cross,red] {};
		\draw (19/16+4,7/16) node[cross,red] {};
		\draw (1/16+4,43/16) node[cross,red] {};
		\draw (30/16+4,30/16) node[cross,red] {};
		\draw (22/16+4,2/16) node[cross,red] {};
		\draw (45/16+4,47.5/16) node[cross,red] {};
		\draw (8/16+4,26/16) node[cross,red] {};
		\draw (7/16+4,7/16) node[cross,red] {};
		\draw (24/16+4,16/16) node[cross,red] {};
		\draw (32/16+4,19/16) node[cross,red] {};
		\draw (39/16+4,28/16) node[cross,red] {};
		\draw (2/16+4,45/16) node[cross,red] {};
		\draw (18/16+4,2/16) node[cross,red] {};
		\draw (42/16+4,8/16) node[cross,red] {};
		\draw (33/16+4,1/16) node[cross,red] {};
		\draw (24/16+4,40/16) node[cross,red] {};
		\draw (19/16+4,46/16) node[cross,red] {};
		\draw (30/16+4,36/16) node[cross,red] {};
		\draw (2/16+4,22/16) node[cross,red] {};
		\draw (45/16+4,11/16) node[cross,red] {};
		\draw (15/16+4,40/16) node[cross,red] {};
		\draw (29/16+4,39/16) node[cross,red] {};
		\draw (11/16+4,45/16) node[cross,red] {};
		\draw (34/16+4,7/16) node[cross,red] {};
		\draw (40/16+4,40/16) node[cross,red] {};
		\draw (39/16+4,44/16) node[cross,red] {};
		\draw (38/16+4,39/16) node[cross,red] {};
		\draw (16/16+4,29/16) node[cross,red] {};
		\draw (12/16+4,33/16) node[cross,red] {};
		\draw (18/16+4,24/16) node[cross,red] {};
		\draw (15/16+4,15/16) node[cross,red] {};
		\draw (22/16+4,36/16) node[cross,red] {};
		\draw (24/16+4,40/16) node[cross,red] {};
		\draw (5/16+4,30/16) node[cross,red] {};
		\draw (8/16+4,18/16) node[cross,red] {};
		\draw (7/16+4,32/16) node[cross,red] {};
		\draw (30/16+4,10/16) node[cross,red] {};
		\draw (36/16+4,43/16) node[cross,red] {};
		\draw (16/16+4,8/16) node[cross,red] {};
		\draw (6/16+4,10/16) node[cross,red] {};
		\draw (26/16+4,12/16) node[cross,red] {};
		\draw (45/16+4,23.5/16) node[cross,red] {};
		\draw (46/16+4,20.5/16) node[cross,red] {};
		\draw (47/16+4,19.5/16) node[cross,red] {};
		\draw (47.4/16+4,16.5/16) node[cross,red] {};
		\draw (47.2/16+4,12.5/16) node[cross,red] {};
		\draw (46.4/16+4,8.5/16) node[cross,red] {};
		\draw (45.3/16+4,1.5/16) node[cross,red] {};
		\draw (45.9/16+4,0.5/16) node[cross,red] {};
		\draw (46.2/16+4,8.5/16) node[cross,red] {};
		\draw (45/16+4,22/16) node[cross,red] {};
		\draw (45.8/16+4,6/16) node[cross,red] {};
		\draw (36/16+4,42.1/16) node[cross,red] {};
		\draw (35.5/16+4,44.9/16) node[cross,red] {};
		\draw (32/16+4,42.5/16) node[cross,red] {};
		\draw (31/16+4,42.9/16) node[cross,red] {};
		\draw (31.5/16+4,43/16) node[cross,red] {};
		\draw (34.4/16+4,43.9/16) node[cross,red] {};
		\draw (34.6/16+4,44.5/16) node[cross,red] {};
		\draw (29/16+4,44.7/16) node[cross,red] {};
		
		\draw (8,0)--(8,3)--(11,3)--(11,0)--(8,0);
		\draw[style=dashed] (8.75,0)--(8.75,3);
		\draw[style=dashed] (9.5,0)--(9.5,3);
		\draw[style=dashed] (10.25,0)--(10.25,3);
		\draw[style=dashed] (8,1.5)--(11,1.5);	
		\draw[style=dashed] (8,0.75)--(11,0.75);	
		\draw[style=dashed] (8,2.25)--(11,2.25);	
		\draw[fill=blue] (8,0) circle (1.5pt);
		\draw[fill=blue] (4/16+8,33/16) circle (1.5pt);
		\draw[fill=blue] (8/16+8,46/16) circle (1.5pt);
		\draw[fill=blue] (11.5/16+8,19/16) circle (1.5pt);
		\draw[fill=blue] (12.5/16+8,37/16) circle (1.5pt);
		\draw[fill=blue] (16.5/16+8,16.5/16) circle (1.5pt);
		\draw[fill=blue] (19.5/16+8,11/16) circle (1.5pt);
		\draw[fill=blue] (23.5/16+8,28/16) circle (1.5pt);
		\draw[fill=blue] (25.5/16+8,22/16) circle (1.5pt);
		\draw[fill=blue] (29/16+8,41/16) circle (1.5pt);
		\draw[fill=blue] (32/16+8,8/16) circle (1.5pt);
		\draw[fill=blue] (33.5/16+8,31.5/16) circle (1.5pt);
		\draw[fill=blue] (37/16+8,26/16) circle (1.5pt);
		\draw[fill=blue] (40.5/16+8,14.5/16) circle (1.5pt);
		\draw[fill=blue] (42.2/16+8,3.5/16) circle (1.5pt);
		\draw[fill=blue] (46.5/16+8,43.5/16) circle (1.5pt);
		\draw (0/16+8,8/16) node[cross,red] {};
		\draw (10/16+8,18/16) node[cross,red] {};
		\draw (43/16+8,22/16) node[cross,red] {};
		\draw (19/16+8,7/16) node[cross,red] {};
		\draw (1/16+8,43/16) node[cross,red] {};
		\draw (30/16+8,30/16) node[cross,red] {};
		\draw (22/16+8,2/16) node[cross,red] {};
		\draw (45/16+8,47.5/16) node[cross,red] {};
		\draw (8/16+8,26/16) node[cross,red] {};
		\draw (7/16+8,7/16) node[cross,red] {};
		\draw (24/16+8,16/16) node[cross,red] {};
		\draw (32/16+8,19/16) node[cross,red] {};
		\draw (39/16+8,28/16) node[cross,red] {};
		\draw (2/16+8,45/16) node[cross,red] {};
		\draw (18/16+8,2/16) node[cross,red] {};
		\draw (42/16+8,8/16) node[cross,red] {};
		\draw (33/16+8,1/16) node[cross,red] {};
		\draw (24/16+8,40/16) node[cross,red] {};
		\draw (19/16+8,46/16) node[cross,red] {};
		\draw (30/16+8,36/16) node[cross,red] {};
		\draw (2/16+8,22/16) node[cross,red] {};
		\draw (45/16+8,11/16) node[cross,red] {};
		\draw (15/16+8,40/16) node[cross,red] {};
		\draw (29/16+8,39/16) node[cross,red] {};
		\draw (11/16+8,45/16) node[cross,red] {};
		\draw (34/16+8,7/16) node[cross,red] {};
		\draw (40/16+8,40/16) node[cross,red] {};
		\draw (39/16+8,44/16) node[cross,red] {};
		\draw (38/16+8,39/16) node[cross,red] {};
		\draw (16/16+8,29/16) node[cross,red] {};
		\draw (12/16+8,33/16) node[cross,red] {};
		\draw (18/16+8,24/16) node[cross,red] {};
		\draw (15/16+8,15/16) node[cross,red] {};
		\draw (22/16+8,36/16) node[cross,red] {};
		\draw (24/16+8,40/16) node[cross,red] {};
		\draw (5/16+8,30/16) node[cross,red] {};
		\draw (8/16+8,18/16) node[cross,red] {};
		\draw (7/16+8,32/16) node[cross,red] {};
		\draw (30/16+8,10/16) node[cross,red] {};
		\draw (36/16+8,43/16) node[cross,red] {};
		\draw (16/16+8,8/16) node[cross,red] {};
		\draw (6/16+8,10/16) node[cross,red] {};
		\draw (26/16+8,12/16) node[cross,red] {};
		\draw (45/16+8,23.5/16) node[cross,red] {};
		\draw (46/16+8,20.5/16) node[cross,red] {};
		\draw (47/16+8,19.5/16) node[cross,red] {};
		\draw (47.4/16+8,16.5/16) node[cross,red] {};
		\draw (47.2/16+8,12.5/16) node[cross,red] {};
		\draw (46.4/16+8,8.5/16) node[cross,red] {};
		\draw (45.3/16+8,1.5/16) node[cross,red] {};
		\draw (45.9/16+8,0.5/16) node[cross,red] {};
		\draw (46.2/16+8,8.5/16) node[cross,red] {};
		\draw (45/16+8,22/16) node[cross,red] {};
		\draw (45.8/16+8,6/16) node[cross,red] {};
		\draw (36/16+8,42.1/16) node[cross,red] {};
		\draw (35.5/16+8,44.9/16) node[cross,red] {};
		\draw (32/16+8,42.5/16) node[cross,red] {};
		\draw (31/16+8,42.9/16) node[cross,red] {};
		\draw (31.5/16+8,43/16) node[cross,red] {};
		\draw (34.4/16+8,43.9/16) node[cross,red] {};
		\draw (34.6/16+8,44.5/16) node[cross,red] {};
		\draw (29/16+8,44.7/16) node[cross,red] {};
		
		\draw (2,-3.5)--(5,-3.5)--(5,-0.5)--(2,-0.5)--(2,-3.5);
		\draw[style=dashed] (2,-2)--(5,-2);
		\draw[style=dashed] (2,-1.25)--(5,-1.25);
		\draw[style=dashed] (2,-2.75)--(5,-2.75);
		\draw[style=dashed] (2,-2.75/2-3.5/2)--(5,-2.75/2-3.5/2);
		\draw[style=dashed] (2,-2.75/2-2/2)--(5,-2.75/2-2/2);
		\draw[style=dashed] (2,-1.25/2-2/2)--(5,-1.25/2-2/2);
		\draw[style=dashed] (2,-1.25/2-0.5/2)--(5,-1.25/2-0.5/2);
		\draw[style=dashed] (3.5,-0.5)--(3.5,-3.5);
		\draw[fill=blue] (2,-3.5) circle (1.5pt);
		\draw[fill=blue] (4/16+2,33/16-3.5) circle (1.5pt);
		\draw[fill=blue] (8/16+2,46/16-3.5) circle (1.5pt);
		\draw[fill=blue] (11.5/16+2,19/16-3.5) circle (1.5pt);
		\draw[fill=blue] (12.5/16+2,37/16-3.5) circle (1.5pt);
		\draw[fill=blue] (16.5/16+2,16.5/16-3.5) circle (1.5pt);
		\draw[fill=blue] (19.5/16+2,11/16-3.5) circle (1.5pt);
		\draw[fill=blue] (23.5/16+2,28/16-3.5) circle (1.5pt);
		\draw[fill=blue] (25.5/16+2,22/16-3.5) circle (1.5pt);
		\draw[fill=blue] (29/16+2,41/16-3.5) circle (1.5pt);
		\draw[fill=blue] (32/16+2,8/16-3.5) circle (1.5pt);
		\draw[fill=blue] (33.5/16+2,31.5/16-3.5) circle (1.5pt);
		\draw[fill=blue] (37/16+2,26/16-3.5) circle (1.5pt);
		\draw[fill=blue] (40.5/16+2,14.5/16-3.5) circle (1.5pt);
		\draw[fill=blue] (42.2/16+2,3.5/16-3.5) circle (1.5pt);
		\draw[fill=blue] (46.5/16+2,43.5/16-3.5) circle (1.5pt);
		\draw (0/16+2,8/16-3.5) node[cross,red] {};
		\draw (10/16+2,18/16-3.5) node[cross,red] {};
		\draw (43/16+2,22/16-3.5) node[cross,red] {};
		\draw (19/16+2,7/16-3.5) node[cross,red] {};
		\draw (1/16+2,43/16-3.5) node[cross,red] {};
		\draw (30/16+2,30/16-3.5) node[cross,red] {};
		\draw (22/16+2,2/16-3.5) node[cross,red] {};
		\draw (45/16+2,47.5/16-3.5) node[cross,red] {};
		\draw (8/16+2,26/16-3.5) node[cross,red] {};
		\draw (7/16+2,7/16-3.5) node[cross,red] {};
		\draw (24/16+2,16/16-3.5) node[cross,red] {};
		\draw (32/16+2,19/16-3.5) node[cross,red] {};
		\draw (39/16+2,28/16-3.5) node[cross,red] {};
		\draw (2/16+2,45/16-3.5) node[cross,red] {};
		\draw (18/16+2,2/16-3.5) node[cross,red] {};
		\draw (42/16+2,8/16-3.5) node[cross,red] {};
		\draw (33/16+2,1/16-3.5) node[cross,red] {};
		\draw (24/16+2,40/16-3.5) node[cross,red] {};
		\draw (19/16+2,46/16-3.5) node[cross,red] {};
		\draw (30/16+2,36/16-3.5) node[cross,red] {};
		\draw (2/16+2,22/16-3.5) node[cross,red] {};
		\draw (45/16+2,11/16-3.5) node[cross,red] {};
		\draw (15/16+2,40/16-3.5) node[cross,red] {};
		\draw (29/16+2,39/16-3.5) node[cross,red] {};
		\draw (11/16+2,45/16-3.5) node[cross,red] {};
		\draw (34/16+2,7/16-3.5) node[cross,red] {};
		\draw (40/16+2,40/16-3.5) node[cross,red] {};
		\draw (39/16+2,44/16-3.5) node[cross,red] {};
		\draw (38/16+2,39/16-3.5) node[cross,red] {};
		\draw (16/16+2,29/16-3.5) node[cross,red] {};
		\draw (12/16+2,33/16-3.5) node[cross,red] {};
		\draw (18/16+2,24/16-3.5) node[cross,red] {};
		\draw (15/16+2,15/16-3.5) node[cross,red] {};
		\draw (22/16+2,36/16-3.5) node[cross,red] {};
		\draw (24/16+2,40/16-3.5) node[cross,red] {};
		\draw (5/16+2,30/16-3.5) node[cross,red] {};
		\draw (8/16+2,18/16-3.5) node[cross,red] {};
		\draw (7/16+2,32/16-3.5) node[cross,red] {};
		\draw (30/16+2,10/16-3.5) node[cross,red] {};
		\draw (36/16+2,43/16-3.5) node[cross,red] {};
		\draw (16/16+2,8/16-3.5) node[cross,red] {};
		\draw (6/16+2,10/16-3.5) node[cross,red] {};
		\draw (26/16+2,12/16-3.5) node[cross,red] {};
		\draw (45/16+2,23.5/16-3.5) node[cross,red] {};
		\draw (46/16+2,20.5/16-3.5) node[cross,red] {};
		\draw (47/16+2,19.5/16-3.5) node[cross,red] {};
		\draw (47.4/16+2,16.5/16-3.5) node[cross,red] {};
		\draw (47.2/16+2,12.5/16-3.5) node[cross,red] {};
		\draw (46.4/16+2,8.5/16-3.5) node[cross,red] {};
		\draw (45.3/16+2,1.5/16-3.5) node[cross,red] {};
		\draw (45.9/16+2,0.5/16-3.5) node[cross,red] {};
		\draw (46.2/16+2,8.5/16-3.5) node[cross,red] {};
		\draw (45/16+2,22/16-3.5) node[cross,red] {};
		\draw (45.8/16+2,6/16-3.5) node[cross,red] {};
		\draw (36/16+2,42.1/16-3.5) node[cross,red] {};
		\draw (35.5/16+2,44.9/16-3.5) node[cross,red] {};
		\draw (32/16+2,42.5/16-3.5) node[cross,red] {};
		\draw (31/16+2,42.9/16-3.5) node[cross,red] {};
		\draw (31.5/16+2,43/16-3.5) node[cross,red] {};
		\draw (34.4/16+2,43.9/16-3.5) node[cross,red] {};
		\draw (34.6/16+2,44.5/16-3.5) node[cross,red] {};
		\draw (29/16+2,44.7/16-3.5) node[cross,red] {};
		
		\draw (6,-3.5)--(9,-3.5)--(9,-0.5)--(6,-0.5)--(6,-3.5);	
		\draw[style=dashed] (6,-2)--(9,-2);
		\draw[style=dashed] (6,-1.25)--(9,-1.25);
		\draw[style=dashed] (6,-2.75)--(9,-2.75);
		\draw[style=dashed] (6,-2.75/2-3.5/2)--(9,-2.75/2-3.5/2);
		\draw[style=dashed] (6,-2.75/2-2/2)--(9,-2.75/2-2/2);
		\draw[style=dashed] (6,-1.25/2-2/2)--(9,-1.25/2-2/2);
		\draw[style=dashed] (6,-1.25/2-0.5/2)--(9,-1.25/2-0.5/2);	
		\draw[style=dashed] (6,-0.5-3/16)--(9,-0.5-3/16);
		\draw[style=dashed] (6,-0.5-9/16)--(9,-0.5-9/16);
		\draw[style=dashed] (6,-0.5-15/16)--(9,-0.5-15/16);
		\draw[style=dashed] (6,-0.5-21/16)--(9,-0.5-21/16);
		\draw[style=dashed] (6,-0.5-27/16)--(9,-0.5-27/16);
		\draw[style=dashed] (6,-0.5-33/16)--(9,-0.5-33/16);
		\draw[style=dashed] (6,-0.5-39/16)--(9,-0.5-39/16);
		\draw[style=dashed] (6,-0.5-45/16)--(9,-0.5-45/16);
		\draw[fill=blue] (6,-3.5) circle (1.5pt);
		\draw[fill=blue] (4/16+6,33/16-3.5) circle (1.5pt);
		\draw[fill=blue] (8/16+6,46/16-3.5) circle (1.5pt);
		\draw[fill=blue] (11.5/16+6,19/16-3.5) circle (1.5pt);
		\draw[fill=blue] (12.5/16+6,37/16-3.5) circle (1.5pt);
		\draw[fill=blue] (16.5/16+6,16.5/16-3.5) circle (1.5pt);
		\draw[fill=blue] (19.5/16+6,11/16-3.5) circle (1.5pt);
		\draw[fill=blue] (23.5/16+6,28/16-3.5) circle (1.5pt);
		\draw[fill=blue] (25.5/16+6,22/16-3.5) circle (1.5pt);
		\draw[fill=blue] (29/16+6,41/16-3.5) circle (1.5pt);
		\draw[fill=blue] (32/16+6,8/16-3.5) circle (1.5pt);
		\draw[fill=blue] (33.5/16+6,31.5/16-3.5) circle (1.5pt);
		\draw[fill=blue] (37/16+6,26/16-3.5) circle (1.5pt);
		\draw[fill=blue] (40.5/16+6,14.5/16-3.5) circle (1.5pt);
		\draw[fill=blue] (42.2/16+6,3.5/16-3.5) circle (1.5pt);
		\draw[fill=blue] (46.5/16+6,43.5/16-3.5) circle (1.5pt);
		\draw (0/16+6,8/16-3.5) node[cross,red] {};
		\draw (10/16+6,18/16-3.5) node[cross,red] {};
		\draw (43/16+6,22/16-3.5) node[cross,red] {};
		\draw (19/16+6,7/16-3.5) node[cross,red] {};
		\draw (1/16+6,43/16-3.5) node[cross,red] {};
		\draw (30/16+6,30/16-3.5) node[cross,red] {};
		\draw (22/16+6,2/16-3.5) node[cross,red] {};
		\draw (45/16+6,47.5/16-3.5) node[cross,red] {};
		\draw (8/16+6,26/16-3.5) node[cross,red] {};
		\draw (7/16+6,7/16-3.5) node[cross,red] {};
		\draw (24/16+6,16/16-3.5) node[cross,red] {};
		\draw (32/16+6,19/16-3.5) node[cross,red] {};
		\draw (39/16+6,28/16-3.5) node[cross,red] {};
		\draw (2/16+6,45/16-3.5) node[cross,red] {};
		\draw (18/16+6,2/16-3.5) node[cross,red] {};
		\draw (42/16+6,8/16-3.5) node[cross,red] {};
		\draw (33/16+6,1/16-3.5) node[cross,red] {};
		\draw (24/16+6,40/16-3.5) node[cross,red] {};
		\draw (19/16+6,46/16-3.5) node[cross,red] {};
		\draw (30/16+6,36/16-3.5) node[cross,red] {};
		\draw (2/16+6,22/16-3.5) node[cross,red] {};
		\draw (45/16+6,11/16-3.5) node[cross,red] {};
		\draw (15/16+6,40/16-3.5) node[cross,red] {};
		\draw (29/16+6,39/16-3.5) node[cross,red] {};
		\draw (11/16+6,45/16-3.5) node[cross,red] {};
		\draw (34/16+6,7/16-3.5) node[cross,red] {};
		\draw (40/16+6,40/16-3.5) node[cross,red] {};
		\draw (39/16+6,44/16-3.5) node[cross,red] {};
		\draw (38/16+6,39/16-3.5) node[cross,red] {};
		\draw (16/16+6,29/16-3.5) node[cross,red] {};
		\draw (12/16+6,33/16-3.5) node[cross,red] {};
		\draw (18/16+6,24/16-3.5) node[cross,red] {};
		\draw (15/16+6,15/16-3.5) node[cross,red] {};
		\draw (22/16+6,36/16-3.5) node[cross,red] {};
		\draw (24/16+6,40/16-3.5) node[cross,red] {};
		\draw (5/16+6,30/16-3.5) node[cross,red] {};
		\draw (8/16+6,18/16-3.5) node[cross,red] {};
		\draw (7/16+6,32/16-3.5) node[cross,red] {};
		\draw (30/16+6,10/16-3.5) node[cross,red] {};
		\draw (36/16+6,43/16-3.5) node[cross,red] {};
		\draw (16/16+6,8/16-3.5) node[cross,red] {};
		\draw (6/16+6,10/16-3.5) node[cross,red] {};
		\draw (26/16+6,12/16-3.5) node[cross,red] {};
		\draw (45/16+6,23.5/16-3.5) node[cross,red] {};
		\draw (46/16+6,20.5/16-3.5) node[cross,red] {};
		\draw (47/16+6,19.5/16-3.5) node[cross,red] {};
		\draw (47.4/16+6,16.5/16-3.5) node[cross,red] {};
		\draw (47.2/16+6,12.5/16-3.5) node[cross,red] {};
		\draw (46.4/16+6,8.5/16-3.5) node[cross,red] {};
		\draw (45.3/16+6,1.5/16-3.5) node[cross,red] {};
		\draw (45.9/16+6,0.5/16-3.5) node[cross,red] {};
		\draw (46.2/16+6,8.5/16-3.5) node[cross,red] {};
		\draw (45/16+6,22/16-3.5) node[cross,red] {};
		\draw (45.8/16+6,6/16-3.5) node[cross,red] {};
		\draw (36/16+6,42.1/16-3.5) node[cross,red] {};
		\draw (35.5/16+6,44.9/16-3.5) node[cross,red] {};
		\draw (32/16+6,42.5/16-3.5) node[cross,red] {};
		\draw (31/16+6,42.9/16-3.5) node[cross,red] {};
		\draw (31.5/16+6,43/16-3.5) node[cross,red] {};
		\draw (34.4/16+6,43.9/16-3.5) node[cross,red] {};
		\draw (34.6/16+6,44.5/16-3.5) node[cross,red] {};
		\draw (29/16+6,44.7/16-3.5) node[cross,red] {};
	\end{tikzpicture}
	\caption{$(0,4,2)$-net in base $2$ (blue points) with point set $\mathcal{X}$ (red crosses)}
	\label{tms-Gewichte}
\end{figure}
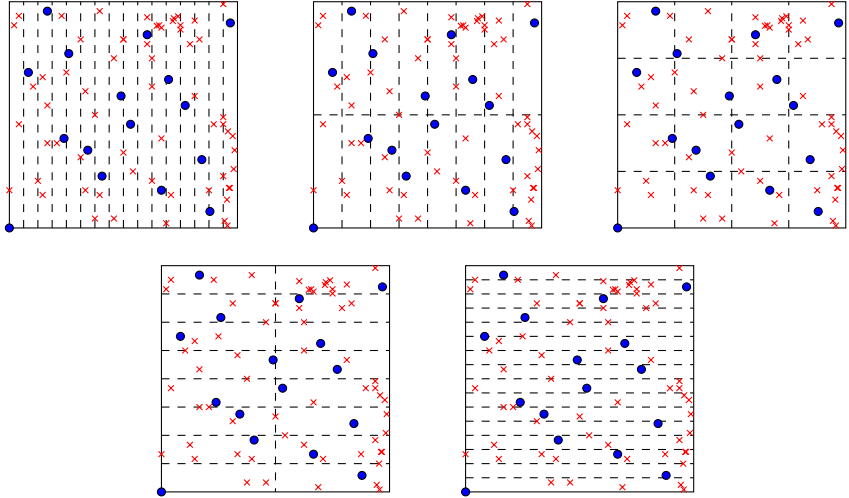

Figure \ref{tms-Gewichte} is a modified version of Figure \ref{042-Netz}. The red crosses represent $\mathcal{X}$, the blue points $P$. For a given blue point, the ratio of points in $\mathcal{X}$ to points in $P$ for that elementary interval is calculated, which contains the blue point. It is necessary to consider all possible partitions, as the ratio in question can vary considerably depending on the partition. This is illustrated in Figure \ref{tms-Gewichte}. 

\subsubsection{Construction of digital nets}\label{Construction of digital nets}

In this subsection, we will give a brief outline for the construction of the aforementioned $(t_{\alpha},m,s)$-nets. If constructed by the following methodology going back to Niederreiter's work \cite{DigitalNets}, they are referred to as digital nets.  We define the finite field with $b$ elements, denoted by $\mathbb{Z}_{b}$, as the set $\{0,1, \ldots, b-1\}$. The addition and multiplication operations are performed modulo $b$ for which we use the notation $x\, (\bmod\ b)$.

We start with the construction of $(t_{1},m,s)$-nets. Let $C_{1}, \ldots, C_{s} \in \mathbb{Z}_{b}^{m \times m}$ be quadratic matrices of size $m$, which will determine the $(t_{1},m,s)$-net. 
Then the $j$-th component $z_{l,j}$ of the $l$-th point of the net is given as follows:
Assume that the $b$-adic expansion of $l-1$ is given by $\sum_{i=0}^{m-1} \lambda_{i} b^{i}$ and define the vector $\vec{\lambda}:= \left(\lambda_{0}, \ldots, \lambda_{m-1}\right)^{\intercal} \in \mathbb{Z}_{b}^{m}$. Furthermore, let the vector $\vec{z}_{l, j}:=\left(z_{l, j, 1},\ldots, z_{l, j, m}\right)^{\intercal}\in \mathbb{Z}_{b}^{m}$ be given by $\vec{z}_{l, j}\equiv C_{j} \vec{\lambda}\, (\bmod\ b)$. The final $z_{l, j}$ then is
\begin{equation*}
z_{l, j}=\frac{z_{l, j, 1}}{b}+\cdots+\frac{z_{l, j, m}}{b^{m}}=\sum_{i=1}^{m}\frac{z_{l, j, i}}{b^{i}}.
\end{equation*}
The determining matrices $C_{1}, \ldots, C_{s}$ must have suitable properties in order to obtain $(t_{\alpha},m,s)$-nets. An illustrative example is the case of the Faure matrices, which are defined as follows:
\begin{equation*}\label{eq:Faure_matrices}
C_{j}:=\left( \begin{array}{ccccc} 
	\binom{1}{1}j^{1-1} & \binom{2}{1}j^{2-1} & \binom{3}{1}j^{3-1} & \cdots & \binom{m}{1}j^{m-1}\\ 
	\binom{1}{2}j^{1-2} & \binom{2}{2}j^{2-2} & \binom{3}{2}j^{3-2} & \cdots & \binom{m}{2}j^{m-2}\\ 
	\vdots & \vdots & \vdots & \vdots & \vdots \\
	\binom{1}{m}j^{1-m} & \binom{2}{m}j^{2-m} & \binom{3}{m}j^{3-m} & \cdots & \binom{m}{m}j^{m-m}\\ 
\end{array} \right) \quad (\bmod\ b).
\end{equation*}
Since $\binom{n}{k}=0$ for $k>n$, these are upper triangular matrices with ones on the diagonal.

\smallskip
\smallskip
Based on the determining matrices of $(t_{1},m,\alpha s)$-nets, we can construct $(t_{\alpha},m,s)$-nets for $\alpha \geq 2$ in the following way. Let $C_{1}, \ldots, C_{\alpha s} \in \mathbb{Z}_{b}^{m \times m}$ be $\alpha s$-many quadratic matrices of size $m$, which determine a $(t_{1},m,\alpha s)$-net. For the $v$-th row of $C_{u}$ we will write $(C_{u})_{v}$, where $u=1,\ldots,\alpha s$ and $v=1,\ldots, m$. We then construct the matrices $D_{j}$, that determine a $(t_{\alpha},m,s)$-net, as 
\begin{equation*}
D_{j}:=\left( \begin{array}{c} (C_{(j-1)\alpha +1})_{1} \\ \vdots \\ (C_{j\alpha })_{1} \\ \vdots \\ (C_{(j-1)\alpha +1})_{m} \\ \vdots \\ (C_{j\alpha })_{m} \end{array} \right), \quad j =1,\ldots,s.
\end{equation*}
Similarly to the previous case, for each integer $l \in\left\{1, \ldots, b^{m}\right\}$, we define the vector $\vec{z}_{l, j}:=D_{j} \vec{\lambda}\, (\bmod\ b)$, with $\vec{\lambda}:= \left(\lambda_{0}, \ldots, \lambda_{m-1}\right)^{\intercal} \in \mathbb{Z}_{b}^{ m}$ consisting of the coefficients of the $b$-adic expansion $l-1=\sum_{i=0}^{m-1} \lambda_{i} b^{i}$. Using $\vec{z}_{l, j}=\left(z_{l, j, 1},\ldots, z_{l, j, \alpha m}\right)^{\intercal}\in \mathbb{Z}_{b}^{m}$, we finally obtain
\begin{equation*}
z_{l, j}=\frac{z_{l, j, 1}}{b}+\cdots+\frac{z_{l, j, \alpha m}}{b^{\alpha m}}=\sum_{i=1}^{\alpha m}\frac{z_{l, j, i}}{b^{i}}.
\end{equation*}

In the following we will use $(t_{1},m,s)$-nets generated by Sobol or Niederreiter-Xing matrices \cite{MagicPointShop,MagicPointShop_webpage}.

\subsubsection{Error bounds}\label{ConvergenceDigitalNets}

While the formulas \eqref{weights_1} and \eqref{weights_2} for the weights look computationally heavy, an advantage of the QMC-averaging method is the availability of a rigorous (determinstic) error bound.

To state these bounds, we need to introduce the following norms:

\begin{definition}[p.12,\ p.14,\ \cite{1}]\label{norm} 
\begin{itemize}
    \item[(i)] For a continuous function $g: [0,1]^{s} \rightarrow \mathbb{R}$, which is $\lambda_s$-almost everywhere  once partially differentiable with respect to each component, we define
$$
\|g\|:= \sum_{\boldsymbol{u} \subseteq\{1,\ldots,s\}} \int_{[0,1]^{|\boldsymbol{u}|}}\left|\partial_{\boldsymbol{z}_{\boldsymbol{u}}} g\left(\boldsymbol{z}_{\boldsymbol{u}}, \mathbf{1}_{-u}\right)\right|\ \mathrm{d} \boldsymbol{z}_{\boldsymbol{u}}.
$$
Here, $\partial_{\boldsymbol{z}_{\boldsymbol{u}}}g\left(\boldsymbol{z}_{\boldsymbol{u}}, \mathbf{1}_{-\boldsymbol{u}}\right)$ denotes the partial mixed derivative of order one for components in $\boldsymbol{u}$. The vector $\left(\boldsymbol{z}_{\boldsymbol{u}}, \boldsymbol{1}_{-\boldsymbol{u}}\right)$ for an arbitrary $\boldsymbol{u} \subseteq\{1, \ldots, s\}$ is $\boldsymbol{z}_{j}$ in its $j$-th component, if $j \in \boldsymbol{u}$, and 1 otherwise.
\item[(ii)] Let $\alpha \geq 2$ and $1 \leq p < \infty$ be integer. The $p,\alpha$-norm of a function $g: [0,1]^{s} \rightarrow \mathbb{R}$, which is $\alpha$-times partially differentiable with respect to each component, is given by
	\begin{small}
		$$
		\begin{aligned}
			\norm{g}_{p, \alpha}^p:= & \sum_{\boldsymbol{u} \subseteq\{1, \ldots, s\}} \sum_{\boldsymbol{v} \subseteq \boldsymbol{u}} \sum_{\boldsymbol{\tau} \in\{1, \ldots, \alpha-1\}^{|\boldsymbol{u} \backslash \boldsymbol{v}|}} \int_{[0,1]^{\left\lvert \boldsymbol{v} \right\rvert}} \left\lvert \int_{[0,1]^{s-\left\lvert \boldsymbol{v} \right\rvert}} \left(\prod_{j \in \boldsymbol{v}} \partial_{z_{j}}^{\alpha} \prod_{j \in \boldsymbol{u} \backslash \boldsymbol{v}} \partial_{z_{j}}^{\tau_{j}}\right) g\left(\boldsymbol{z}\right) d \boldsymbol{z}_{\{1, \ldots, s\} \backslash \boldsymbol{v}}\right\rvert^{p} d \boldsymbol{z}_{v}.
		\end{aligned}
		$$
	\end{small}
	Here, $\left(\prod_{j \in \boldsymbol{v}} \partial_{z_{j}}^{\alpha} \prod_{j \in \boldsymbol{u} \backslash \boldsymbol{v}} \partial_{z_{j}}^{\tau_{j}}\right) g(\boldsymbol{z})$ denotes the partial mixed derivative of order $\alpha$ or $\tau_{j}$ in the $j$-th coordinate.
\end{itemize}
\end{definition}

Both norms appear naturally when dealing with Quasi-Monte Carlo integration, see e.g. \cite{Niederreiter,Walsh}. 

\smallskip

In \cite{1} several error bounds are given.  Their proofs are based on 
   the Koksma-Hlawka inequality \eqref{Koksma-Hlawka}  and on a Walsh-series analysis (Appendix A in \cite{Walsh}). Optimizing the choice of $\nu$ with respect to $m$ and $t_{\alpha}$ yields the following bounds (see Section 4.2 in  \cite{1}) :

\begin{theorem}[Corollary 12,\ \cite{1}]
	Let $P=\left\{\boldsymbol{z}_{1}, \ldots, \boldsymbol{z}_{L}\right\}$ be a digital $(t_{1},m,s)$-net in base $b$ with $L=b^m$ and  let $\nu = \frac{m}{2}$. Moreover, define $y_{\max} := \max_{n=1,\ldots,N}  \left\lvert y_{n} \right\rvert $.
 Then, there exists a constant $C_{s,b,t_{1},y_{\max}} >0$ such that
\begin{equation}\label{eq:conv_first_order}
	\left|\operatorname{err}\left(f_{\theta}\right)-\operatorname{app}^{avg}_{L}\left(f_{\theta}\right)\right| \leq C_{s,b,t_{1},y_{\max}}\left( \left\|f_{\theta}^{2}\right\| + \left\|f_{\theta}\right\| + \left\|f_{\theta}^{2}\right\|_{2,2}  +\left\|f_{\theta}\right\|_{2,2} \right) \log_{b}(L)^{2s-1} L^{-\frac{1}{2}}.
\end{equation}
\end{theorem}

\smallskip
\smallskip

We now turn our attention to the case in which the order $\alpha$ is at least $2$. 

\begin{theorem}[Corollary 14,\ \cite{1}]\label{KonvOrd2komplett}
	Let $P=\left\{\boldsymbol{z}_{1}, \ldots, \boldsymbol{z}_{L}\right\}$ be a digital $(t_{\alpha}, m, s)$-net in base $b$  of order $\alpha \geq 2$. Moreover, let $L=b^m$ and 
 $\nu = \frac{\alpha}{\alpha+1} m$ and define $y_{\max} := \max_{n=1,\ldots,N}  \left\lvert y_{n} \right\rvert $. Then, there exists a constant  $C_{\alpha,s,b,t_{\alpha},y_{\max}}>0$ such that
\begin{equation}\label{eq:conv_higher_order}
	\left|\operatorname{err}\left(f_{\theta}\right)-\operatorname{app}^{avg}_{L}\left(f_{\theta}\right)\right|\leq C_{\alpha,s,b,t_{\alpha},y_{\max}}\left( \left\|f_{\theta}^{2}\right\|_{2,\alpha}  + \left\|f_{\theta}\right\|_{2,\alpha}  \right) \log_{b}(L)^{\alpha s} L^{-\frac{\alpha}{\alpha+1} } .
\end{equation}
\end{theorem}

The convergence order in the case of $t_1$-nets is similar to Monte Carlo subsampling with order $1/2$, but  the error bound has the advantage of being non-random in comparison to Monte Carlo subsampling. For $\alpha \geq 2$ one even obtains the better convergence order $\alpha/(1+\alpha)$ (up to logarithmic terms).

\subsubsection{Implementation}
 The most complex part of the implementation is clearly to calculate the weights  \eqref{weights_1} and \eqref{weights_2}. Here  the quantities $\left|\mathcal{X}_{\boldsymbol{d}}\left(\boldsymbol{z}_{l}\right)\right|$ and $\sum_{\substack{n=1, \boldsymbol{x}_{n} \in I_{\boldsymbol{d}}\left(\boldsymbol{z}_{l}\right)}}^{N} y_{n}$ have to be computed for each elementary interval with $|\boldsymbol{d}|=\nu-q$. In total, there are $\binom{\nu-q+s-1}{s-1}$ different vectors $\boldsymbol{d}$, that satisfy this property. Thus, for every value of $l$ between $1$ and $b^{m}$  and for each value of $0\leq q\leq \min\left(\nu,s-1\right)$, we need to determine the quantities $S_{r}(\boldsymbol{z}_{l})$ and $T_{r}(\boldsymbol{z}_{l})$ defined by
\begin{equation*} 
S_{r}(\boldsymbol{z}_{l}):= \sum_{\substack{\boldsymbol{d} \in \mathbb{N}_{0}^{s}\\ |\boldsymbol{d}|=r}}\left|\mathcal{X}_{\boldsymbol{d}}\left(\boldsymbol{z}_{l}\right)\right|, \quad T_{r}(\boldsymbol{z}_{l}):= \sum_{\substack{\boldsymbol{d} \in \mathbb{N}_{0}^{s} \\|\boldsymbol{d}|=r}} \sum_{n=1}^{N} y_{n} \mathbbm{1}_{I_{\boldsymbol{d}}\left(\boldsymbol{z}_{l}\right)} \left(\boldsymbol{x}_{n}\right),
\end{equation*}
where $r=\nu-q$. In fact, if $y_n=1$ for $n=1, \ldots, N$, then the sum of indicator functions $\sum_{n=1}^{N} \mathbbm{1}_{I_{\boldsymbol{d}}\left(\boldsymbol{z}_{l}\right)}$ counts the number of original data points, which are in the same elementary interval as $\boldsymbol{z}_{l}$ and we obtain $S_r(\boldsymbol{z})$.

The following algorithm computes $T_{r}(\boldsymbol{z}_{l})$. In order to obtain $S_{r}(\boldsymbol{z}_{l})$, one just sets $y_{n} = 1$ for $n = 1, \ldots, N$.

\newpage

\hrule
\smallskip
\noindent{\textbf{Algorithm (W)}}: Calculation of $S_{r}(\boldsymbol{z})$ and $T_{r}(\boldsymbol{z})$ \cite{1}
\smallskip
\hrule 
\smallskip
\noindent{\textbf{input}}: $\boldsymbol{z}\in [0,1)^{s}$, $\mathcal{X}$, $\mathcal{Y}$, $b$, $r\geq0$\\
\textbf{output}: $T_{r}(\boldsymbol{z})$ (or $S_{r}(\boldsymbol{z})$ if $y_n=1$ for $n=1:N$)
\begin{lstlisting}[escapeinside={(*}{*)}]
set (*$T_{r}\left(\boldsymbol{z}\right)=0$*)
for (*$n$*) = 1:(*$N$*)
    for (*$j$*)=1:(*$s$*)
	find maximal (*$i_{j} \in\{0, \ldots, r\}$*), so the first (*$i_{j}$*) digits of the (*$b$*)-adic 
	expansion of (*$z_{j}$*) and (*$\left(\boldsymbol{x}_{n}\right)_{j}$*) are equal
    end
    set (*$\boldsymbol{i}=\left(i_{1}, \ldots, i_{s}\right)$*) and (*$T_{r}(\boldsymbol{z})=T_{r}(\boldsymbol{z})+y_{n}\#\left\{\boldsymbol{d} \in \mathbb{N}_{0}^{s}:|\boldsymbol{d}|=r, \boldsymbol{d} \leq \boldsymbol{i}\right\}$*)
end
\end{lstlisting}
\hrule

\smallskip
\smallskip
\smallskip

The choice $i_{j}=0$ in line 4 will occur if none of the coefficients of the $b$-adic expansions of $z_j$ and $(x_n)_j$ are identical. The objective of Algorithm (W) is to identify the smallest $s$-dimensional elementary interval, which has a length of at least $b^{-r}$ for each one-dimensional elementary interval contained within it, and which includes $\boldsymbol{z}$. This is accomplished for each data point $\boldsymbol{x}_{n}$. Given a specific $\boldsymbol{x}_{n}$, we count the number of vectors $\boldsymbol{d}\in \mathbb{N}_{0}^{s}$ with $|\boldsymbol{d}|=r$, such that $\boldsymbol{x}_{n}$ is part of $I_{\boldsymbol{d}}\left(\boldsymbol{z}_{l}\right)$.
The numbers $N_{r, \boldsymbol{i}}:=\#\left\{\boldsymbol{d} \in \mathbb{N}_{0}^{s}:|\boldsymbol{d}|=r, \boldsymbol{d} \leq \boldsymbol{i}\right\}$ are universal, i.e.~independent of the data and regression function, and can be computed in advance. See \cite{1} for a particular algorithm for this task, which leads to a total cost of $\mathcal{O}(r s N)$ for Algorithm (W) with a  storage space of order $\mathcal{O}(\max(s,r))$.

\smallskip
\smallskip

In order to calculate the weights, the following equation must be used:
\begin{equation*}
W_{\mathcal{X}, \mathcal{Y}, P, \nu, l}=\frac{b^{\nu-m}}{N} \sum_{q=0}^{\min(s-1,\nu)}(-1)^{q}\binom{s-1}{q} \frac{1}{b^{q}} T_{\nu-q}(\boldsymbol{z}_{l}).
\end{equation*}
Choosing $ S_{\nu -q}(\boldsymbol{z}_{l})$ instead of $T_{\nu-q}(\boldsymbol{z}_{l})$ results in the weights $W_{\mathcal{X}, P, \nu, l}$. The cost of computing all weights is of order $\mathcal{O}(b^{m}ms^{2}N)$. 

As we mentioned in the beginning, the weights are independent of $\theta$. Therefore, if $\mathcal{X}$, $\mathcal{Y}$ and $P$ remain constant throughout the optimization process, it is sufficient to calculate the weights once at the outset. The procedure for fitting the neural network  merely requires updating
\begin{equation} \label{eq:app_L-chap:3}
\operatorname{app}_{L}^{avg}(f_{\theta})=\sum_{l=0}^{L-1} f_{\theta}^{2}\left(\boldsymbol{z}_{l}\right) W_{\mathcal{X}, P, \nu, l}-2 \sum_{l=0}^{L-1} f_{\theta}\left(\boldsymbol{z}_{l}\right) W_{\mathcal{X}, \mathcal{Y}, P, \nu, l}+\frac{1}{N} \sum_{n=1}^{N} y_{n}^{2}
\end{equation}
in each optimization step. Assuming that $f_{\theta}^{2}$ and $f_{\theta}$ can be evaluated with cost $\mathcal{O}(s)$, we end up with cost of order $\mathcal{O}(Ls)$, rather than $\mathcal{O}(Ns)$, which is the cost if we use $\operatorname{err}\left(f_{\theta}\right)$. We require additional storage of $\mathcal{O}(L)$ for the weights and $\sum_{n=1}^{N} y_{n}^{2}$. This suggests that training the neural network with the compressed data set will be faster. However, it is not yet clear how much time will be saved in the whole optimization procedure, due to the additional effort required at the outset.

\subsection{Supercompress method}

A classical approach to deal with regression problems are $K$-nearest 
 neighbors algorithms, which in their simplest form estimate the regression function at a point $\boldsymbol{z}$ by averaging the responses $y_{i}$ of the $K$ data points $\boldsymbol{x}_{i}$, which are the $K$ nearest points to $\boldsymbol{z}$. See e.g. \cite{devroye} Chapters 2 and 14 in \cite{Hastie}. These algorithms are not  data reduction methods themselves. However, data reduction can be achieved by incorporating clustering, which leads, e.g. to  the supercompress method proposed by \cite{supercompress}.

The aim of the supercompress method is again to find a compressed point set $P=\left\{\boldsymbol{z}_{1},\ldots, \boldsymbol{z}_{K}\right\}$, but  instead of weights, one obtains the corresponding (approximate) responses $\mathcal{W} = \{w_{1},\ldots,w_{K}\}$.  This is achieved through a  specific $K$-means clustering, which does not  employ a conventional clustering approach on the input space, but utilizes the responses $\mathcal{Y}$. 
More precisely, the supercompress method aims to
find the points $\left\{\boldsymbol{z}_{1},\ldots, \boldsymbol{z}_{K}\right\}$ such that the loss function
\begin{equation}\label{eq:supercompress_objective}
L = \sum_{j=1}^{K} L_{j} = \sum_{j=1}^{K} \sum_{i \in I_{j}} \left( y_{i}-w_{j}\right)^{2}
\end{equation}
with the approximate responses \begin{equation} w_{j} = \frac{1}{\left|I_{j}\right|}\sum_{i \in I_{j}} y_{i}, \quad j=1,\ldots,K, \end{equation} and the Voronoi-clusters
\begin{equation}
I_{j}= \left\{i\in \{1,\ldots,N\}:\left\|\boldsymbol{z}_{j}-\boldsymbol{x}_{i}\right\|_{2} \leq \left\|\boldsymbol{z}_{j^{\prime}}-\boldsymbol{x}_{i}\right\|_{2} \text { for all } j^{\prime} \neq j\right\}, \quad j=1, \ldots, K,
\end{equation} is minimized. In contrast, the classical $K$-means algorithm on the input space would find the  data points $\left\{\boldsymbol{z}'_{1},\ldots, \boldsymbol{z}'_{K}\right\}$ and Voronoi-clusters $I_1', \ldots, I_K'$, which minimize the loss 
\begin{equation*}
L' = \sum_{j=1}^{K} L_{j}' = \sum_{j=1}^{K} \sum_{i \in I_{j}'} \left\| \boldsymbol{z}'_{i}-\boldsymbol{x}_{i}\right \|^{2}.
\end{equation*}
Although clustering on the input space is already known to be a NP-hard problem, see e.g. \cite{NP}, many fast algorithm exist, which converge at least to a local minimum. See e.g. \cite{fastCluster}.
For practical implementation of the supercompress method, the authors of \cite{supercompress}  propose the following 
iterative  $K$-means algorithm:

\smallskip
\smallskip

\hrule
\smallskip
\noindent{\textbf{Algorithm (S)}}: Calculation of supercompress data compression \cite{supercompress}
\smallskip
\hrule 
\smallskip
\noindent{\textbf{input}}: $\mathcal{X}$, $\mathcal{Y}$, $K$\\
\textbf{output}: $P = \{\boldsymbol{z}_{1},\ldots,\boldsymbol{z}_{K}\}$, $\mathcal{W}= \{w_{1},\ldots,w_{K}\}$
\begin{lstlisting}[escapeinside={(*}{*)}]
set (*$P = \{\}$*), (*$\mathcal{I} = \{\}$*) and (*$\mathcal{W} = \{\}$*)
split (*$\mathcal{X}$*) into 2 (*$K$*)-means clusters with centers (*$\{\boldsymbol{z}_{1},\boldsymbol{z}_{2}\}$*) and partitions (*$\{I_{1},I_{2}\}$*)
for (*$j=1,2$*) compute responses (*$w_{j} = \frac{1}{\left|I_{j}\right|}\sum_{i \in I_{j}} y_{i}$*) and losses (*$L_{j} = \sum_{i \in I_{j}} \left( y_{i}-w_{j}\right)^{2}$*)
set (*$P = P \cup \{\boldsymbol{z}_{1},\boldsymbol{z}_{2}\}$*), (*$\mathcal{I} = \mathcal{I} \cup \{I_{1},I_{2}\}$*) and (*$\mathcal{W} = \mathcal{W} \cup \{w_{1},w_{2}\}$*)
for (*$k$*) = 3:(*$K$*)
    find index (*$j^{\star} = \max_{j=1,\ldots,k-1} L_{j}$*)
    split cluster (*$j^{\star}$*) into 2 (*$K$*)-means clusters with centers (*$\{\boldsymbol{z}_{j^{\star}},\boldsymbol{z}_{k}\}$*) and 
    partitions (*$\{I_{j^{\star}},I_{k}\}$*) 
    for (*$j=j^{\star},k$*) update responses (*$w_{j} = \frac{1}{\left|I_{j}\right|}\sum_{i \in I_{j}} y_{i}$*) and losses (*$L_{j} = \sum_{i \in I_{j}} \left( y_{i}-w_{j}\right)^{2}$*)
    set (*$P = P \cup \{\boldsymbol{z}_{k}\}$*), (*$\mathcal{I} = \mathcal{I} \cup \{I_{k}\}$*) and (*$\mathcal{W} = \mathcal{W} \cup \{w_{k}\}$*) and update cluster (*$j^{\star}$*)
end
\end{lstlisting}
\hrule

\smallskip
\smallskip
\smallskip

The initial step involves partitioning the input space $\mathcal{X}$ into two clusters using a $K$-means approach. The loss for each cluster is then calculated according to the formula $L_{j} = \sum_{i \in I_{j}} \left( y_{i}-w_{j}\right)^{2}$. The cluster with the higher loss is divided into two clusters based on $K$-means on the input space once more. This results in two new clusters, each with a center and a corresponding response. These two new clusters replace the old cluster with the highest loss. The losses $L_{j}$ for the two new clusters are updated. Then, the cluster with the greatest loss is divided into two clusters in the same way as before. This procedure is repeated until $K$ clusters have been obtained. Each cluster has a center point, denoted by $\boldsymbol{z}_{k}$, which is part of the compressed data, and a corresponding response, denoted by $w_{k}$, which is contained in $\mathcal{W}$. 

\smallskip
\smallskip
\smallskip
In \cite{supercompress} the authors propose another, more robust algorithm by taking the convex combination 
\begin{equation}\label{eq:robust_objective}
\widetilde{L}=\lambda \sum_{j=1}^{K} \sum_{i \in I_{j}} \left\| \boldsymbol{z}_{j}-\boldsymbol{x}_{i}\right\|_{2}^{2}+(1-\lambda) \sum_{j=1}^{K} \sum_{i \in I_{j}} \left( y_{i}-w_{j}\right)^{2}.
\end{equation}
Here $\lambda \in[0,1]$ is a weight parameter quantifying the trade-off between the two terms. When $\lambda=0$, we return to the case of the original supercompress algorithm. If $\lambda=1$, the problem reduces to the traditional $K$-means problem. One can interpret this modified criterion as a trade-off between a fully supervised reduction strategy $(\lambda=0)$, which fully incorporates response information for data reduction, and a fully robust reduction strategy $(\lambda=1)$, which reduces the data using only the input feature information.
In particular, a default choice of $\lambda=1/(s+1)$ is suggested in \cite{supercompress}. This is motivated by the observation that the objective function $\widetilde{L}$ then becomes proportional to
\begin{equation*}
\sum_{j=1}^{K} \sum_{i \in I_{j}} \frac{1}{s} \left\| \boldsymbol{z}_{j}-\boldsymbol{x}_{i}\right\|_{2}^{2}+\sum_{j=1}^{K} \sum_{i \in I_{j}} \left( y_{i}-w_{j}\right)^{2}.
\end{equation*}

In order to optimize the new loss function $\widetilde{L}$, it is possible to use Algorithm (S) with a slight modification. This involves changing the $y_{i}$'s to $\tilde{y}_{i}=y_{i}\sqrt{(1-\lambda) / \lambda} $, and exchanging the loss functions $L_{j}$ with
\begin{equation*}
\widetilde{L}_{j}=\sum_{i \in I_{j}}\left ( \left\|\boldsymbol{z}_{j}-\boldsymbol{x}_{i}\right\|_{2}^{2}+\left(\tilde{y}_{i}-w_{j}\right)^{2}\right ).
\end{equation*}
We will refer to this more robust algorithm, which optimizes $\widetilde{L}$, as robust supercompress.

Error estimates for this supercompress method in terms of the number $K$ of compressed points seem to be unknown. One can consider the standard $K$-means clustering as a sub-optimal solution  of a discrete quantization problem. For the latter the error of the optimal quantization decays as $\mathfrak{o}(K^{-1/d})$. See e.g. Chapter II.6 in \cite{graf_luschgy}. Thus, we can expect at best a similar behavior for the approximation of \begin{equation*}
\operatorname{err}\left(f_{\theta}\right):=\frac{1}{N} \sum_{n=1}^{N}\left(f_{\theta}\left(\boldsymbol{x}_{n}\right)-y_{n}\right)^{2}
\end{equation*}
by
\begin{equation*}
	\operatorname{app}_{L}^{clst}(f_{\theta}):= \frac{1}{K}\sum_{k=1}^{K} \left(f_{\theta}\left(\boldsymbol{z}_{k}\right)-w_{k}\right)^{2}.
\end{equation*}

\subsection{QMC-Voronoi method}

While testing the Quasi-Monte Carlo compression approach, we observed several phenomena. To understand their source, in particular  whether they are caused by the weights or the digital nets,
we tried a method which combines (unsupervised) clustering and the use of digital nets. We call this method QMC-Voronoi method.

To construct the compressed data set for the QMC-Voronoi method, we start with a digital net. Instead of calculating complex weights, we form clusters by using the Voronoi diagram, see e.g.~Section I.1 in \cite{graf_luschgy}, on the digital net. This implies that for each data point $\boldsymbol{x}$, the closest point $\boldsymbol{z}$ of the digital net $P$ is identified based on the Euclidean distance, and $\boldsymbol{x}$ is assigned to the Voronoi region of $\boldsymbol{z}$. Hence, $P$ represents the compressed set. For each compressed point $\boldsymbol{z}$, the corresponding response $w$ is found by taking the average of the corresponding responses of the data points $\boldsymbol{x}$, which are in the Voronoi region of $\boldsymbol{z}$. Consequently, as for the supercompress method the error $\operatorname{err}(f_{\theta})$ can therefore be approximated by $\operatorname{app}_{L}^{clst}(f_{\theta})$. The above procedure is summarized in Algorithm (V), where we take $K=b^m$ points for the compressed data set.

This method can be seen as a specific clustering on the input space $\mathcal{X}$ with prescribed centers of the clusters, so we expect a behavior of the error similar or worse than  for the (robust) supercompression method.

\newpage

\hrule
\smallskip
\noindent{\textbf{Algorithm (V)}}: Calculation of QMC-Voronoi data compression
\smallskip
\hrule 
\smallskip
\noindent{\textbf{input}}: $\mathcal{X}$, $\mathcal{Y}$, $b$, $m$\\
\textbf{output}: $P \subseteq \{\boldsymbol{z}_{1},\ldots,\boldsymbol{z}_{K}\}$, $\mathcal{W}= \{w_{1},\ldots,w_{K}\}$
\begin{lstlisting}[escapeinside={(*}{*)}]
create digital net (*$P = \{\boldsymbol{z}_{1},\ldots,\boldsymbol{z}_{K}\}$*)
for (*$i$*) = 1:(*$N$*)
    find closest point (*$\boldsymbol{z} \in P$*) to (*$\boldsymbol{x}_{i}$*) and asign (*$y_{i}$*) to (*$\boldsymbol{z}$*)
end
for (*$k$*) = 1:(*$K$*)
    if at least one (*$y_{i}$*) got asigned to (*$\boldsymbol{z}_{k}$*)
        compute (*$w_{k}$*) as the average of each response (*$y_{i}$*), which is asigned to (*$\boldsymbol{z}_{k}$*)
    else
        delete (*$\boldsymbol{z}_{k}$*) from (*$P$*) 
    end
end
\end{lstlisting}
\hrule

\smallskip
\smallskip

\section{Numerical results}
\label{sec:3}

Finally, we undertake a numerical comparison of the methods. We start by considering three fixed functions $f$ and compare $\operatorname{err}(f)$  with $\operatorname{app}_{L}^{avg}(f_{\theta})$ and $\operatorname{app}_{L}^{clst}(f_{\theta})$, respectively.
Subsequently, we train a neural network using the MNIST dataset \cite{MNIST} and compare the accuracy achieved with the different compression methods. All implementations can be found in our Git repository \cite{Git}.

\subsection{Test functions}\label{subsec:test_functions}
The following functions are taken from the virtual library \cite{TestFunctions}.
For an input point $\boldsymbol{x} = \left(x_{1},\ldots,x_{s}\right) \in [0,1)^{s}$ and vectors $\boldsymbol{a} = \left(a_{1},\ldots,a_{s}\right) \in \mathbb{R}^{s}$ and $\boldsymbol{u} = \left(u_{1},\ldots,u_{s}\right) \in [0,1)^{s}$ we define 
\begin{equation}\label{continuous integrand}
	f_1\left( \boldsymbol{x}\right) := \exp\left(-\sum_{i=1}^{s} a_{i} \left| x_{i}-u_{i} \right| \right).
\end{equation}
This function satisfies $\|f_1\|< \infty $ as well as $V(f_1)< \infty$, but has no higher regularity w.r.t. $\|\cdot\|_{p.\alpha}$ for $\alpha \geq 2$, due to the involved absolute values.
 The second test function is given by
\begin{equation}\label{discontinuous integrand}
	f_2\left( \boldsymbol{x}\right) := 
	\begin{cases}
		0 & \text{if } x_{1}>u_{1} \text{ or } x_{2}>u_{2}, \\
		\exp\left(\sum_{i=1}^{s} a_{i} x_{i} \right) & \text{otherwise}.
	\end{cases}
\end{equation} This function is discontinuous, so it has no regularity w.r.t. $\|\cdot\|_{p,\alpha}$ and $\|\cdot\|$, but still has finite Hardy-Krause variation, i.e. $V(f_2)< \infty$.
Finally, the so-called Zhou function is given by
\begin{equation}\label{Zhou}
	f_3\left( \boldsymbol{x}\right) := \frac{10^{s}}{2} \left(\phi\left(10\boldsymbol{x}-\frac{10}{3} \right)  + \phi\left(10\boldsymbol{x}-\frac{20}{3}\right) \right),
\end{equation}
where
$$  \phi\left(\boldsymbol{x}\right) = \frac{1}{\left(2\pi\right)^{\frac{s}{2}}} \exp\left(-\frac{\left\|\boldsymbol{x}\right\|^{2}}{2}\right).$$
This function satisfies  $V(f_3)< \infty$, $\|f_3\|< \infty $ as well as $\|f_3\|_{p,\alpha} < \infty$ for any $p\geq 1$ and $\alpha \geq 2$.

Note that these functions have different scales. For example, for $s=2$ the second function attains values between zero and $\max\{ 1, \exp(a_1u_1)\} \max\{ 1, \exp(a_2u_2)\}$, while the first one takes values between zero and one only. Finally, Zhou's function contains values between zero and  at most $7.9579$. In the following we will choose $u_{1}=\ldots=u_{s}=0.5$ so that $f_1$ is symmetric around the center of the unit cube and the discontinuity of $f_2$ appears at the same position. Furthermore we set $a_{1}=\ldots=a_{s}=5$. Thus the maximum value of $f_2$ for $s=2$ is $(\exp(2.5))^2 \approx 148.41$, which is much larger than that of $f_1$ and $f_3$.
For more information and illustrations of these functions, see \cite{TestFunctions}.

For evaluating the different compression methods, a total of $N=3000$ uniformly distributed points are sampled in the interval $[0,1)^{s}$. For each point, the function is evaluated and an independent  perturbation from the $\mathcal{N}(0,0.02)$-distribution is added. Subsequently, the compressed point set $P$ is generated with $L=K$ points, where $L$ is selected from the set $\{2^{5},2^{6},2^{7},2^{8},2^{9},2^{10}\}$. Finally, in Table \ref{tab:1} we compare the approximation error $\left| \operatorname{err}(f) -\operatorname{app}_{K}^{clst}(f) \right|$ of the supercompress method to the one of the QMC-averaging method, i.e.~$\left| \operatorname{err}(f) - \operatorname{app}_{L}^{avg}(f) \right|$, for different compression rates $K/N$. This is repeated 100 times and the observed values are averaged in order to enhance the robustness of the result.
For the QMC-averaging method we use the "Magic Point Shop" \cite{MagicPointShop,MagicPointShop_webpage} to construct the set $P$. In fact, we utilize Niederreiter-Xing matrices of order 1, as presented in \cite{MagicPointShop,MagicPointShop_webpage}. 

\begin{table}[h!]
	\begin{center}
		\newcommand{\resultrow}[7]{#1 & #2 & #3 & #4 & #6 & #5 & #7\\}
		\begin{tabular}{>{\centering}m{1.9cm}|m{0.65cm}|m{0.65cm}|m{0.6cm}|m{2cm}|m{2cm}|m{2cm}|}
			\multicolumn{1}{c|}{compression rate} & \multicolumn{1}{c}{$N$} & \multicolumn{1}{c}{$s$} & \multicolumn{1}{c|}{$K$} & \multicolumn{1}{c}{$f_1$} & \multicolumn{1}{c}{$f_2$} & \multicolumn{1}{c}{$f_3$}\\
			\cmidrule(r){1-7}
			\multicolumn{1}{c|}{\multirow{2}{*}{$1 \%$}} & \multicolumn{1}{c}{\multirow{2}{*}{$3000$}} & \multicolumn{1}{c}{\multirow{2}{*}{$2$}}  & \multicolumn{1}{c|}{\multirow{2}{*}{$2^{5}$}} & \multicolumn{1}{c}{$3,1054\cdot10^{-4}$}  & \multicolumn{1}{c}{$2,2135\cdot 10^{4}$}  & \multicolumn{1}{c}{$4,86\cdot10^{-2}$}\\ & \multicolumn{1}{c}{} & \multicolumn{1}{c}{} & \multicolumn{1}{c|}{} & \multicolumn{1}{c}{$2,24\cdot10^{-2}$}  & \multicolumn{1}{c}{$4,4712\cdot 10^{4}$}  & \multicolumn{1}{c}{$3,5859\cdot10^{0}$}\\
			\cmidrule(r){1-7}
			\multicolumn{1}{c|}{\multirow{2}{*}{$4 \%$}} & \multicolumn{1}{c}{\multirow{2}{*}{$3000$}} & \multicolumn{1}{c}{\multirow{2}{*}{$2$}}  & \multicolumn{1}{c|}{\multirow{2}{*}{$2^{7}$}} & \multicolumn{1}{c}{$3,3239\cdot10^{-4}$}  & \multicolumn{1}{c}{$1,2331\cdot 10^{3}$}  & \multicolumn{1}{c}{$2,5\cdot10^{-3}$}\\ & \multicolumn{1}{c}{} & \multicolumn{1}{c}{} & \multicolumn{1}{c|}{} & \multicolumn{1}{c}{$1,37\cdot10^{-2}$}  & \multicolumn{1}{c}{$2,0453\cdot 10^{4}$}  & \multicolumn{1}{c}{$2,0974\cdot10^{0}$}\\
			\cmidrule(r){1-7}
			\multicolumn{1}{c|}{\multirow{2}{*}{$9 \%$}} & \multicolumn{1}{c}{\multirow{2}{*}{$3000$}} & \multicolumn{1}{c}{\multirow{2}{*}{$2$}}  & \multicolumn{1}{c|}{\multirow{2}{*}{$2^{8}$}} & \multicolumn{1}{c}{$2,5366\cdot10^{-4}$}  & \multicolumn{1}{c}{$2,9351\cdot 10^{2}$}  & \multicolumn{1}{c}{$5,98\cdot10^{-4}$}\\ & \multicolumn{1}{c}{} & \multicolumn{1}{c}{} & \multicolumn{1}{c|}{} & \multicolumn{1}{c}{$1,9\cdot10^{-3}$}  & \multicolumn{1}{c}{$6,9552\cdot 10^{3}$}  & \multicolumn{1}{c}{$1,1437\cdot10^{0}$}\\
			\cmidrule(r){1-7}
			\multicolumn{1}{c|}{\multirow{2}{*}{$17 \%$}} & \multicolumn{1}{c}{\multirow{2}{*}{$3000$}} & \multicolumn{1}{c}{\multirow{2}{*}{$2$}}  & \multicolumn{1}{c|}{\multirow{2}{*}{$2^{9}$}} & \multicolumn{1}{c}{$1,7051\cdot10^{-4}$}  & \multicolumn{1}{c}{$8,7618\cdot 10^{1}$}  & \multicolumn{1}{c}{$1,2786\cdot10^{-4}$}\\ & \multicolumn{1}{c}{} & \multicolumn{1}{c}{} & \multicolumn{1}{c|}{} & \multicolumn{1}{c}{$1,9\cdot10^{-3}$}  & \multicolumn{1}{c}{$6,4545 \cdot 10^{3}$}  & \multicolumn{1}{c}{$1,1702\cdot10^{0}$}\\
			\cmidrule(r){1-7}
			\cmidrule(r){1-7}
			\multicolumn{1}{c|}{\multirow{2}{*}{$34 \%$}} & \multicolumn{1}{c}{\multirow{2}{*}{$3000$}} & \multicolumn{1}{c}{\multirow{2}{*}{$2$}}  & \multicolumn{1}{c|}{\multirow{2}{*}{$2^{10}$}} & \multicolumn{1}{c}{$1,0473\cdot10^{-4}$}  & \multicolumn{1}{c}{$1,9970 \cdot 10^{1}$}  & \multicolumn{1}{c}{$1,5783\cdot10^{-5}$}\\ & \multicolumn{1}{c}{} & \multicolumn{1}{c}{} & \multicolumn{1}{c|}{} & \multicolumn{1}{c}{$2\cdot10^{-3}$}  & \multicolumn{1}{c}{$8,2950\cdot 10^{3}$}  & \multicolumn{1}{c}{$4,878\cdot10^{-1}$}\\
			\cmidrule(r){1-7}
		\end{tabular}
		\caption{average error (top row supercompress, bottom row QMC-averaging)}
		\label{tab:1}
	\end{center}
\end{table}

Table \ref{tab:1} illustrates the results for the dimension $s=2$. The top row expresses the average error of the supercompress algorithm, while the bottom row refers to that of the QMC-averaging method. It can be observed that the supercompress algorithm performs better for each function. Only for the discontinuous function $f_2$ and high compression rates, i.e. small $K$, both methods perform equally bad.
Since $f_2([0,1]^2)=[0, \exp(5)]$, and $\exp(5) \approx 148,4132$, the observed behavior might be a consequence of the lack of scaling.
To analyze this, we repeated the procedure for $\gamma f_2 / \| f_2 \|_{\infty}$ with different scales $\gamma$.  In Table \ref{tab:2} we see, that both methods yield a smaller error for $\gamma=1$, which corresponds to standardized test functions. Thus, both methods behave similar for all test functions when the latter are properly scaled; the supercompress method seems to be always the superior choice. Interestingly, the regularity of the test functions does not seem to have a big influence on the performance of the QMC-averaging method.  You can find a scaled version with $\gamma=1$ of all tables regarding the error analysis in the Appendix (Table \ref{tab:1_scaled}, Table \ref{tab:3_scaled}, Table \ref{tab:4_scaled}).

\begin{table}[h!]
	\begin{center}
		\begin{tabular}{>{\centering}m{1.9cm}|m{0.65cm}|m{0.65cm}|m{0.6cm}|m{2.5cm}|m{2.5cm}|}
			\multicolumn{1}{c|}{compression rate} & \multicolumn{1}{c}{$N$} & \multicolumn{1}{c}{$s$} & \multicolumn{1}{c|}{$K$} & \multicolumn{1}{c}{no scale} & \multicolumn{1}{c}{scale 1}\\
			\cmidrule(r){1-6}
			\multicolumn{1}{c|}{\multirow{2}{*}{$9 \%$}} & \multicolumn{1}{c}{\multirow{2}{*}{$3000$}} & \multicolumn{1}{c}{\multirow{2}{*}{$2$}}  & \multicolumn{1}{c|}{\multirow{2}{*}{$2^{8}$}} & \multicolumn{1}{c}{$2,9351\cdot 10^{2}$} & \multicolumn{1}{c}{$1,3326\cdot 10^{-2}$}\\ & \multicolumn{1}{c}{}  & \multicolumn{1}{c}{} & \multicolumn{1}{c|}{} & \multicolumn{1}{c}{$6,9552\cdot 10^{3}$} & \multicolumn{1}{c}{$3,1577\cdot 10^{-1}$} \\
			\cmidrule(r){1-6}
			\multicolumn{1}{c|}{\multirow{2}{*}{$17 \%$}} & \multicolumn{1}{c}{\multirow{2}{*}{$3000$}} & \multicolumn{1}{c}{\multirow{2}{*}{$2$}}  & \multicolumn{1}{c|}{\multirow{2}{*}{$2^{9}$}} & \multicolumn{1}{c}{$8,7618\cdot 10^{1}$} & \multicolumn{1}{c}{$3,9779\cdot 10^{-3}$}\\ & \multicolumn{1}{c}{} & \multicolumn{1}{c}{} & \multicolumn{1}{c|}{} & \multicolumn{1}{c}{$6,4545\cdot 10^{3}$} & \multicolumn{1}{c}{$2,9304 \cdot 10^{-1}$}\\
			\cmidrule(r){1-6}
		\end{tabular}
		\caption{average error for $f_2$ for differently scaled function values}
		\label{tab:2}
	\end{center}
\end{table}

We also performed the same numerical experiment without adding noisy perturbations. The results are found to be consistent with those above, indicating that adding noise does not significantly influence the comparison. In fact, as long as the noise level of the data is reasonably low, the effects of the above analysis remain the same.

The sensitivity of both methods in terms of the dimension is presented in Table \ref{tab:3}. For $f_1$, the size of the problem does not affect the accuracy of our methods. However, it can be observed that there is a deterioration in accuracy for the other functions. In particular, the error increases significantly in higher dimensions when considering $f_2$. Note that for the largest dimension $s=10$, QMC-averaging leads to smaller errors for $f_1$ and $f_3$ than supercompress. We do not have a mathematical explanation at hand and did not explore this phenomenon further due to the very high running time of QMC-averaging for larger $s$; see also Table \ref{tab:7}.
\begin{table}[h!]
	\begin{center}
		\begin{tabular}{>{\centering}m{1.9cm}|m{0.65cm}|m{0.65cm}|m{0.6cm}|m{2cm}|m{2cm}|m{2cm}|}
			\multicolumn{1}{c|}{compression rate} & \multicolumn{1}{c}{$N$} & \multicolumn{1}{c}{$s$} & \multicolumn{1}{c|}{$K$} & \multicolumn{1}{c}{$f_1$} & \multicolumn{1}{c}{$f_2$} & \multicolumn{1}{c}{$f_3$}\\
			\cmidrule(r){1-7}
			\multicolumn{1}{c|}{\multirow{2}{*}{$9 \%$}} & \multicolumn{1}{c}{\multirow{2}{*}{$3000$}} & \multicolumn{1}{c}{\multirow{2}{*}{$2$}}  & \multicolumn{1}{c|}{\multirow{2}{*}{$2^{8}$}} & \multicolumn{1}{c}{$2,5366\cdot10^{-4}$}  & \multicolumn{1}{c}{$2,9352\cdot 10^{2}$}  & \multicolumn{1}{c}{$5,98\cdot10^{-4}$}\\ & \multicolumn{1}{c}{} & \multicolumn{1}{c}{} & \multicolumn{1}{c|}{} & \multicolumn{1}{c}{$1,9\cdot10^{-3}$}  & \multicolumn{1}{c}{$6,9552\cdot 10^{3}$}  & \multicolumn{1}{c}{$1,1437\cdot10^{0}$}\\
			\cmidrule(r){1-7}
			\multicolumn{1}{c|}{\multirow{2}{*}{$9 \%$}} & \multicolumn{1}{c}{\multirow{2}{*}{$3000$}} & \multicolumn{1}{c}{\multirow{2}{*}{$3$}}  & \multicolumn{1}{c|}{\multirow{2}{*}{$2^{8}$}} & \multicolumn{1}{c}{$2,6782\cdot10^{-4}$}  & \multicolumn{1}{c}{$3,8841 \cdot 10^{6}$}  & \multicolumn{1}{c}{$5,03\cdot10^{-2}$}\\ & \multicolumn{1}{c}{} & \multicolumn{1}{c}{} & \multicolumn{1}{c|}{} & \multicolumn{1}{c}{$1,9\cdot10^{-3}$}  & \multicolumn{1}{c}{$7,2364\cdot 10^{7}$}  & \multicolumn{1}{c}{$7,7694\cdot10^{0}$}\\
			\cmidrule(r){1-7}
			\multicolumn{1}{c|}{\multirow{2}{*}{$9 \%$}} & \multicolumn{1}{c}{\multirow{2}{*}{$3000$}} & \multicolumn{1}{c}{\multirow{2}{*}{$5$}}  & \multicolumn{1}{c|}{\multirow{2}{*}{$2^{8}$}} & \multicolumn{1}{c}{$3,286\cdot10^{-4}$}  & \multicolumn{1}{c}{$4,3730\cdot 10^{13}$}  & \multicolumn{1}{c}{$1,4568\cdot 10^{1}$}\\ & \multicolumn{1}{c}{} & \multicolumn{1}{c}{} & \multicolumn{1}{c|}{} & \multicolumn{1}{c}{$2,3514\cdot10^{-4}$}  & \multicolumn{1}{c}{$6,8260\cdot 10^{14}$}  & \multicolumn{1}{c}{$8,9822\cdot 10^{1}$}\\
			\cmidrule(r){1-7}
			\multicolumn{1}{c|}{\multirow{2}{*}{$9 \%$}} & \multicolumn{1}{c}{\multirow{2}{*}{$3000$}} & \multicolumn{1}{c}{\multirow{2}{*}{$10$}}  & \multicolumn{1}{c|}{\multirow{2}{*}{$2^{8}$}} & \multicolumn{1}{c}{$3,5688\cdot10^{-4}$}  & \multicolumn{1}{c}{$1,0240\cdot 10^{29}$}  & \multicolumn{1}{c}{$2,9730\cdot 10^{5}$}\\ & \multicolumn{1}{c}{} & \multicolumn{1}{c}{} & \multicolumn{1}{c|}{} & \multicolumn{1}{c}{$5,0307\cdot10^{-8}$}  & \multicolumn{1}{c}{$8,2610 \cdot 10^{31}$}  & \multicolumn{1}{c}{$7,4510 \cdot 10^{3}$}\\
			\cmidrule(r){1-7}
		\end{tabular}
		\caption{average error (top row supercompress, bottom row QMC-averaging)}
		\label{tab:3}
	\end{center}
\end{table}


In order to gain further insight into the behavior of QMC-averaging, we designed the QMC-Voronoi method and conducted a comparison between its performance and that of the QMC-averaging method of \cite{1}. The results can be seen in Table \ref{tab:4}. The upper row displays the results for the QMC-averaging method, while the bottom row presents the average error regarding the QMC-Voronoi method. The setup of this experiment is the same as the one for the comparison of QMC-averaging and supercompress.

\begin{table}[h!]
	\begin{center}
		\begin{tabular}{>{\centering}m{1.9cm}|m{0.65cm}|m{0.65cm}|m{0.6cm}|m{2cm}|m{2cm}|m{2cm}|}
			\multicolumn{1}{c|}{compression rate} & \multicolumn{1}{c}{$N$} & \multicolumn{1}{c}{$s$} & \multicolumn{1}{c|}{$K$} & \multicolumn{1}{c}{$f_1$} & \multicolumn{1}{c}{$f_2$} & \multicolumn{1}{c}{$f_3$}\\
			\cmidrule(r){1-7}
			\multicolumn{1}{c|}{\multirow{2}{*}{$4 \%$}} & \multicolumn{1}{c}{\multirow{2}{*}{$3000$}} & \multicolumn{1}{c}{\multirow{2}{*}{$2$}}  & \multicolumn{1}{c|}{\multirow{2}{*}{$2^{7}$}} & \multicolumn{1}{c}{$1,37\cdot10^{-2}$}  & \multicolumn{1}{c}{$2,0453\cdot 10^{4}$}  & \multicolumn{1}{c}{$2,0974\cdot10^{0}$}\\ & \multicolumn{1}{c}{} & \multicolumn{1}{c}{} & \multicolumn{1}{c|}{} & \multicolumn{1}{c}{$4,4655\cdot10^{-4}$}  & \multicolumn{1}{c}{$2,6420\cdot 10^{3}$}  & \multicolumn{1}{c}{$6,92\cdot10^{-2}$}\\
			\cmidrule(r){1-7}
			\multicolumn{1}{c|}{\multirow{2}{*}{$9 \%$}} & \multicolumn{1}{c}{\multirow{2}{*}{$3000$}} & \multicolumn{1}{c}{\multirow{2}{*}{$2$}}  & \multicolumn{1}{c|}{\multirow{2}{*}{$2^{8}$}} & \multicolumn{1}{c}{$1,9\cdot10^{-3}$}  & \multicolumn{1}{c}{$6,9552\cdot 10^{3}$}  & \multicolumn{1}{c}{$1,1437\cdot10^{0}$}\\ & \multicolumn{1}{c}{} & \multicolumn{1}{c}{} & \multicolumn{1}{c|}{} & \multicolumn{1}{c}{$7,0245\cdot10^{-5}$}  & \multicolumn{1}{c}{$4,6737\cdot 10^{2}$}  & \multicolumn{1}{c}{$3,36\cdot10^{-2}$}\\
			\cmidrule(r){1-7}
			\multicolumn{1}{c|}{\multirow{2}{*}{$17 \%$}} & \multicolumn{1}{c}{\multirow{2}{*}{$3000$}} & \multicolumn{1}{c}{\multirow{2}{*}{$2$}}  & \multicolumn{1}{c|}{\multirow{2}{*}{$2^{9}$}} & \multicolumn{1}{c}{$1,9\cdot10^{-3}$}  & \multicolumn{1}{c}{$6,4545\cdot 10^{3}$}  & \multicolumn{1}{c}{$1,1702\cdot10^{0}$}\\ & \multicolumn{1}{c}{} & \multicolumn{1}{c}{} & \multicolumn{1}{c|}{} & \multicolumn{1}{c}{$1,7547\cdot10^{-4}$}  & \multicolumn{1}{c}{$1,3446\cdot 10^{3}$}  & \multicolumn{1}{c}{$2,41\cdot10^{-2}$}\\
			\cmidrule(r){1-7}
		\end{tabular}
		\caption{average error (top row QMC-averaging, bottom row QMC-Voronoi)}
		\label{tab:4}
	\end{center}
\end{table}
The results of QMC-Voronoi consistently outperform those of the QMC-averaging method in all scenarios. In fact, its performance is not too far away from the one of the supercompress method. This indicates that clustering algorithms are more effective for the considered problem.

Additionally, a significant drawback of the QMC-averaging method is its complex calculation. In order to include all partitions of the unit cube into elementary intervals, we use alternating sums with binomial weights, which may result in significant numerical instability. At the same time, we perform an approximation for every partition, which leads to a possible accumulation of errors. The QMC-Voronoi approach considers only one partition of the unit cube, which is more flexible than the one based on elementary intervals.

To underline this observation, we evaluated the running time required to compute the compressed data sets. A total of $N=10000$ data points are sampled uniformly at random from the interval $[0,1)^s$. The corresponding responses are determined by the function $f_2$ given by (\ref{discontinuous integrand}). This procedure is repeated 20 times and the mean is taken. The results are presented in Table \ref{tab:7}. The dimension $s$ of the space is varied in the first column and the number of compressed points $K$ is varied in the first row.

\begin{table}[h!]
	\begin{center}
		\begin{tabular}{>{\centering}m{1cm}|m{2cm}|m{2cm}|m{2cm}|m{2cm}|}
			\multicolumn{1}{c|}{$s\backslash K$} & \multicolumn{1}{c}{$2^6$} & \multicolumn{1}{c}{$2^8$} & \multicolumn{1}{c}{$2^{10}$} & \multicolumn{1}{c}{$2^{12}$} \\
			\cmidrule(r){1-5}
			\multicolumn{1}{c|}{\multirow{3}{*}{$2$}} & \multicolumn{1}{c}{$5,7\cdot10^{-3}$} & \multicolumn{1}{c}{$1,69\cdot10^{-2}$} & \multicolumn{1}{c}{$5,35\cdot10^{-2}$}  & \multicolumn{1}{c}{$1,980\cdot10^{-1}$} \\
			& \multicolumn{1}{c}{$1,813\cdot10^{-1}$} & \multicolumn{1}{c}{$8,122\cdot10^{-1}$} & \multicolumn{1}{c}{$3,2312\cdot10^{0}$}  & \multicolumn{1}{c}{$1,3529\cdot10^{1}$} \\
			& \multicolumn{1}{c}{$5,53\cdot10^{-2}$} & \multicolumn{1}{c}{$6,34\cdot10^{-2}$} & \multicolumn{1}{c}{$7,36\cdot10^{-2}$}  & \multicolumn{1}{c}{$7,97\cdot10^{-2}$} \\
			\cmidrule(r){1-5}
			\multicolumn{1}{c|}{\multirow{3}{*}{$3$}} & \multicolumn{1}{c}{$6,3\cdot10^{-3}$} & \multicolumn{1}{c}{$1,85\cdot10^{-2}$}  & \multicolumn{1}{c}{$5,39\cdot10^{-2}$} & \multicolumn{1}{c}{$1,957\cdot10^{-1}$} \\
			& \multicolumn{1}{c}{$3,257\cdot10^{-1}$} & \multicolumn{1}{c}{$1,5233\cdot10^{0}$} & \multicolumn{1}{c}{$6,1759\cdot10^{0}$}  & \multicolumn{1}{c}{$2,5003\cdot 10^{1}$} \\
			& \multicolumn{1}{c}{$5,47\cdot10^{-2}$} & \multicolumn{1}{c}{$5,87\cdot10^{-2}$} & \multicolumn{1}{c}{$6,49\cdot10^{-2}$}  & \multicolumn{1}{c}{$8,10\cdot10^{-2}$} \\
			\cmidrule(r){1-5}
			\multicolumn{1}{c|}{\multirow{3}{*}{$5$}} & \multicolumn{1}{c}{$5,6\cdot10^{-3}$} & \multicolumn{1}{c}{$1,71\cdot10^{-2}$} & \multicolumn{1}{c}{$5,44\cdot10^{-2}$}  & \multicolumn{1}{c}{$1,979\cdot10^{-1}$} \\
			& \multicolumn{1}{c}{$5,735\cdot10^{-1}$} & \multicolumn{1}{c}{$3,0826\cdot10^{0}$} & \multicolumn{1}{c}{$1,2265\cdot10^1$}  & \multicolumn{1}{c}{$5,3843\cdot 10^{1}$} \\
			& \multicolumn{1}{c}{$5,56\cdot10^{-2}$} & \multicolumn{1}{c}{$6,42\cdot10^{-2}$} & \multicolumn{1}{c}{$7,53\cdot10^{-2}$}  & \multicolumn{1}{c}{$9,23\cdot10^{-2}$} \\
			\cmidrule(r){1-5}
		\end{tabular}
		\caption{average running time for data compression in seconds (top row supercompress, middle row QMC-averaging, bottom row QMC-Voronoi)}
		\label{tab:7}
	\end{center}
\end{table}

Although the supercompress algorithm performs the best, it does not take the most time. In fact, it is the fastest algorithm for low $K$. We observe that the QMC-averaging algorithm takes the most time. As previously indicated, this is caused by the calculation of the weights, which averages over many different partitions. Additionally, the supercompress and QMC-Voronoi algorithms are more robust regarding the dimension. Even for a slight increase in $s$, the QMC-averaging algorithm doubles its running time. For large values of $K$, the QMC-Voronoi method is computed at a faster rate than the supercompress method. Furthermore, the cost of the QMC-Voronoi method does not increase at the same rate as the supercompress method in relation to $K$.

\subsection{Neural networks}

\begin{figure}[h]
	\centering
	\includegraphics{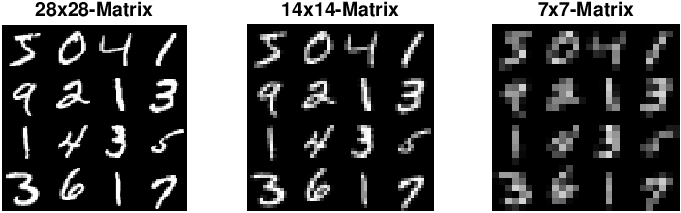}
	\caption{visualization of the precompressed data}
	\label{fig:compression}
\end{figure}

Thus far, we have considered only uniformly random data and a rather simple connection between data points and responses. However, in general this is not the case. To illustrate this, we will now examine the MNIST data set, which can be found in \cite{MNIST}. This data set contains $6\cdot10^4$ grey-scale images of handwritten numbers between $0$ and $9$. Each image is represented by a matrix of dimensions $(28 \times 28)$, with entries from $0$ to $1$. The value of each entry indicates the darkness of the corresponding pixel. By concatenating all the pixels in a given image, we obtain the image of the handwritten number. Consequently, the set $\mathcal{X}$ contains the pixel sets\footnote{Here we identify $1$ as $1-{\tt eps}$, where ${\tt eps}$ is the machine accuracy. Thus, we can still work with  the assumption $\mathcal{X} \subset [0,1)^s$.}, while the set $\mathcal{Y}$ contains the corresponding handwritten numbers. Figure \ref{fig:compression} illustrates 16 examples of such pixel sets. In the case of an entry of $0$, the pixel is depicted as black.

Our aim is to train a neural network with the data in question. To compare the different methods, the accuracies of the trained neural networks will be evaluated on a test data set, which is part of the MNIST data set. 

It should be noted that the dimension of a matrix with dimensions $(28 \times 28)$ is much larger than the dimensions of the points that have been considered thus far. In order to reduce the running time, particularly that required for calculating the weights, we implement a precompression strategy. Initially, we limit our consideration to the first $N=10^{4}$ grey-scale images. Additionally, we reduce the dimension by transforming each $(28 \times 28)$-matrix to a $(14 \times 14)$-matrix. This results in a reduction of the dimension from $s=784$ to $s=196$. The idea is to take a submatrix of a pixel set with dimensions of two by two and represent it by its average value. Figure \ref{fig:compression} illustrates the impact of this precompression on the data. On the left, we have examples of the original data. In the middle, the $(2 \times 2)$-submatrices are exchanged by their average value, while the same process occurs for $(4 \times 4)$-submatrices on the right.

It can be observed that, despite the increased pixelation, the $(14 \times 14)$-matrices can still be read, at least to some extent. Further compression of the data to $(7 \times 7)$-matrices renders it more challenging for the human eye to discern the individual values. This is the reason why we opted to train the network with $(14 \times 14)$-matrices.

\smallskip
\smallskip

After these preparations, the neural network can be trained. We start by compressing the data in accordance with the compression algorithms established in section \ref{DataCompressionMethods}. Afterwards, the neural network is trained with the compressed set. In order to have a broad overview, we will use the QMC-averaging method, the (robust) supercompress method, the QMC-Voronoi method and the traditional $K$-means clustering. In fact, $K$-means and (robust) supercompress appear to be similar in many respects. However, there is a key difference between them in the way they consider the image space. $K$-means clusters solely based on the $\boldsymbol{x}$-space, whereas robust supercompress attempts to combine clustering based on the input space and the $y$-space. The normal supercompress algorithm is focused on finding clusters based on the output space.

In this instance, the dimension of the input data is too large to utilise Niederreiter-Xing matrices. Consequently, Sobol matrices \cite{MagicPointShop,MagicPointShop_webpage} are employed as the construction matrices of the digital net, with $\nu$ set to 2. The design of the neural network and its training are based on \cite{NN} with 100 epochs. For clustering algorithms, this procedure can be employed without modification. The QMC-averaging method requires adjustments to the neural network, as it must be trained based on points and weights instead of points and corresponding responses.

\begin{figure}[h!]
	\centering
	\includegraphics[width=9cm,height=7.1cm]{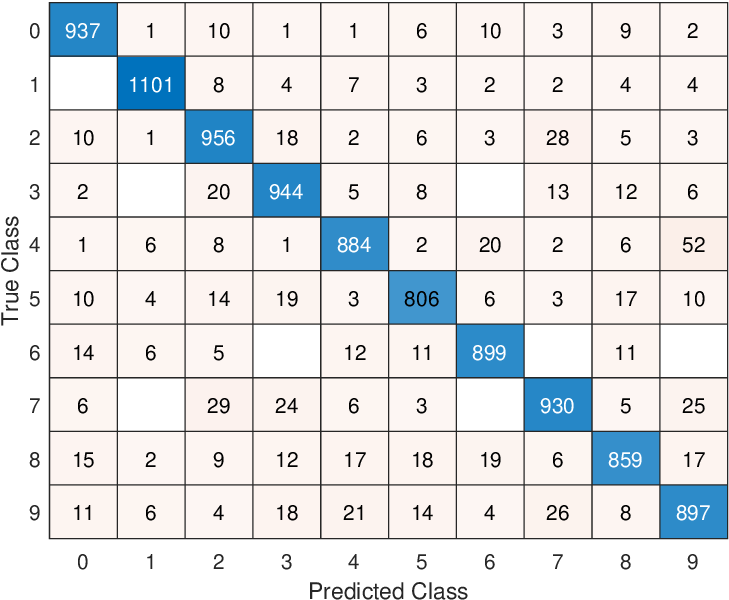}
	\caption{confusion chart of the neural network without compression}
	\label{fig:NN_no_compression}
\end{figure}

At the end of each training, the final loss and the required time (including compression and training) are observed. Additionally, we test the neural network to obtain the accuracy (the relative number of times it predicts the correct number) and a confusion chart. In order to assess the efficacy of the compression method, we initially train the neural network with the uncompressed data. The resulting confusion chart is presented in Figure \ref{fig:NN_no_compression}. 

If our prediction is accurate, the result will be on the diagonal. The number indicates the number of times the neural network predicts the class on the x-axis, while the class on the y-axis is the correct one. For example, 1 is predicted correctly 1101 times. The least accurate combination is 9 and 4. The neural network predicts 52 times a 9, despite the real written number being a 4. It is evident that the objective is to maximize the numbers on the diagonal.

A comparison of the performance of the compression algorithms with that of the uncompressed training is presented in Table \ref{tab:5}. The table shows the accuracy, the required time (i.e. $12:34$ corresponds to 12 minutes and 34 seconds), and the final loss, with the exception of the QMC-averaging method, for which the complex input structure (weights instead of corresponding responses) does not allow a comparable loss value. The number of compressed points $L=K$ is varied for values in the set $\{2^{9},2^{10},2^{11},2^{12}\}$. We have dropped the results of QMC-averaging and $K$-means for $L=K=2^{12}$, due to no significant improvement in accuracy.

\begin{table}[h!]
	\begin{center}
		\begin{tabular}{>{\centering}m{1.9cm}|m{0.65cm}|m{2cm}|m{2cm}|m{2cm}|}
			\multicolumn{1}{c|}{method} & \multicolumn{1}{c|}{compression rate} & \multicolumn{1}{c}{accuracy} & \multicolumn{1}{c}{time [min]} & \multicolumn{1}{c}{loss}\\
			\cmidrule(r){1-5}
			\multicolumn{1}{c|}{\multirow{1}{*}{no compression}} & \multicolumn{1}{c|}{-} & \multicolumn{1}{c}{$92,13 \%$} & \multicolumn{1}{c}{$13:00$} & \multicolumn{1}{c}{0,0215} \\
			\cmidrule(r){1-5}
			\multicolumn{1}{c|}{\multirow{4}{*}{supercompress}} & \multicolumn{1}{c|}{$5 \%$} & \multicolumn{1}{c}{$55,15 \%$} & \multicolumn{1}{c}{$0:57$} & \multicolumn{1}{c}{0,3533} \\
			& \multicolumn{1}{c|}{$10 \%$} & \multicolumn{1}{c}{$72,68 \%$} & \multicolumn{1}{c}{$1:26$} & \multicolumn{1}{c}{0,2799} \\
			& \multicolumn{1}{c|}{$20 \%$} & \multicolumn{1}{c}{$88,16 \%$} & \multicolumn{1}{c}{$2:45$} & \multicolumn{1}{c}{0,164} \\
			& \multicolumn{1}{c|}{$40 \%$} & \multicolumn{1}{c}{$91,98 \%$} & \multicolumn{1}{c}{$6:47$} & \multicolumn{1}{c}{0,1126} \\ 
			\cmidrule(r){1-5}
			\multicolumn{1}{c|}{\multirow{4}{*}{robust supercompress}} & \multicolumn{1}{c|}{$5 \%$} & \multicolumn{1}{c}{$53,45 \%$} & \multicolumn{1}{c}{$0:44$} & \multicolumn{1}{c}{0,2748} \\
            & \multicolumn{1}{c|}{$10 \%$} & \multicolumn{1}{c}{$55,33\%$} & \multicolumn{1}{c}{$1:24$} & \multicolumn{1}{c}{0,2389} \\
            & \multicolumn{1}{c|}{$20 \%$} & \multicolumn{1}{c}{$61,51 \%$} & \multicolumn{1}{c}{$2:46$} & \multicolumn{1}{c}{0,1678} \\
            & \multicolumn{1}{c|}{$40 \%$} & \multicolumn{1}{c}{$64,89 \%$} & \multicolumn{1}{c}{$5:23$} & \multicolumn{1}{c}{0,1165} \\ 
			\cmidrule(r){1-5}
			\multicolumn{1}{c|}{\multirow{3}{*}{QMC}} & \multicolumn{1}{c|}{$5 \%$} & \multicolumn{1}{c}{$8,92 \%$} & \multicolumn{1}{c}{$14:28$} & \multicolumn{1}{c}{-} \\
            & \multicolumn{1}{c|}{$10 \%$} & \multicolumn{1}{c}{$10,28 \%$} & \multicolumn{1}{c}{$29:08$} & \multicolumn{1}{c}{-} \\
            & \multicolumn{1}{c|}{$20 \%$} & \multicolumn{1}{c}{$10,09 \%$} & \multicolumn{1}{c}{$59:02$} & \multicolumn{1}{c}{-} \\ 
			\cmidrule(r){1-5}
			\multicolumn{1}{c|}{\multirow{3}{*}{K-means}} & \multicolumn{1}{c|}{$5 \%$} & \multicolumn{1}{c}{$11,01 \%$} & \multicolumn{1}{c}{$1:57$} & \multicolumn{1}{c}{0,4367} \\
            & \multicolumn{1}{c|}{$10 \%$} & \multicolumn{1}{c}{$9,2 \%$} & \multicolumn{1}{c}{$3:58$} & \multicolumn{1}{c}{0,4457} \\
            & \multicolumn{1}{c|}{$20 \%$} & \multicolumn{1}{c}{$11,22 \%$} & \multicolumn{1}{c}{$7:37$} & \multicolumn{1}{c}{0,4513} \\ 
			\cmidrule(r){1-5}
		\end{tabular}
		\caption{comparison of the different compression methods for the training of a neural network}
		\label{tab:5}
	\end{center}
\end{table}

In general, supercompress performs the best. Its normal version is more accurate than the robust one, though it takes a bit longer. QMC-averaging and $K$-means are very poor. It seems that they are as poor as guessing. Additionally, they take longer, especially QMC-averaging. This is caused by the high effort for precalculating the weights. Consequently, they are not suitable for this compression problem. Why is that so?

We observed a striking phenomenon about the weights. As previously stated in Section \ref{sec:weighted_digital_nets}, the weights serve to indicate the relative importance of each point within the digital net. It is notable that for all compression rates, the weights associated with the first two points (of which one is always the zero vector) are considerably higher than those of the remaining points. Indeed, the latter are largely approximately zero. This implies that the majority of the points are located within the same elementary interval as the first two points for a significant number of partitions. Consequently, the QMC-averaging point set is not an accurate representation of the data set $\mathcal{X}$. This is primarily due to particular structure of $\mathcal{X}$. Since a substantial number of pixels are completely black (have a 0-entry), it is understandable that a large proportion of points is close to the $0$-vector. Therefore, a representative set that is well distributed on the unit cube, is not a good choice for our data. In fact, this behavior can be even more drastically observed when attempting to use the QMC-Voronoi algorithm to train the network. It is noteworthy that all points from the set $\mathcal{X}$ are assigned to only two points (in fact, the first two points) in the set $P$. Consequently, these points are the only points for which a corresponding response exists and, therefore, the only points to train the neural network with. Given that two points are insufficient for training a neural network, this method could not be included in Table \ref{tab:5}.

The reason why $K$-means clustering is not performing well may be attributed to the clustering based on the $\boldsymbol{x}$-space. This causes many clusters to contain points with different corresponding responses. This is not a problem for a cluster like $\{2,2,2,2,2,2,2,4,4\}$, since the rounded cluster mean response is still $2$. Though, clusters such as $\{2,2,2,2,4,4,4,4\}$ correspond to a response of $3$, which is not even a response within the cluster. Therefore incorporating the $y$-space into the clustering method appears to be a logical approach. This could also explain why the robust version performs less effectively in this instance. It incorporates the $\boldsymbol{x}$-space into the loss function instead of just the $y$-space. As a result, for this discrete $y$-space problem, a clustering approach based on the output space, performs the best. One potential solution to this issue is to employ a different rounding method. Instead of taking the average of the responses, we could select the response that appears the most frequently. However, this does not address the fundamental challenge that methods primarily using the $\boldsymbol{x}$-space to cluster can produce clusters with a multitude of corresponding responses.

\smallskip
\smallskip

\begin{figure}[h]
	\centering
	\begin{tabular}{c c}
		\textbf{\ \ \ supercompress} & \textbf{\ \ \ robust supercompress} \\
		\includegraphics[width=7.2cm,height=6cm]{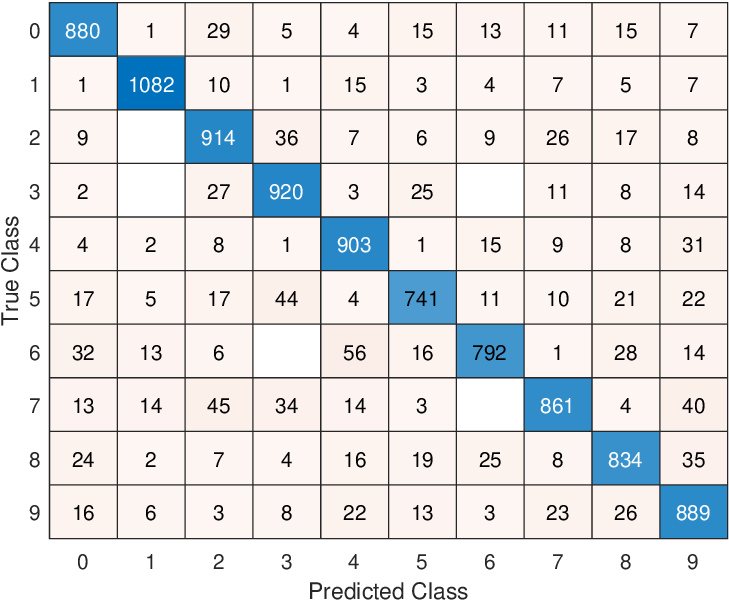} & \includegraphics[width=7.2cm,height=6cm]{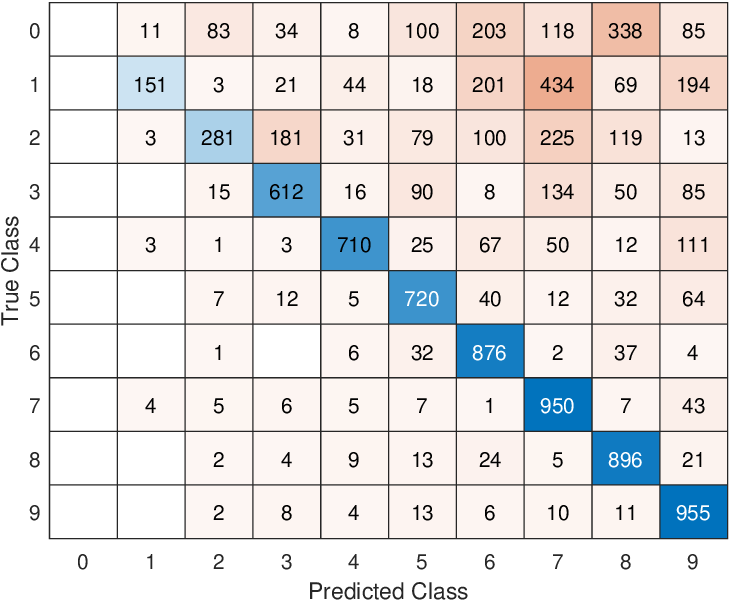}\\
		\textbf{\ \ \ quasi-Monte Carlo} & \textbf{\ \ \ K-means} \\
		\includegraphics[width=7.2cm,height=6cm]{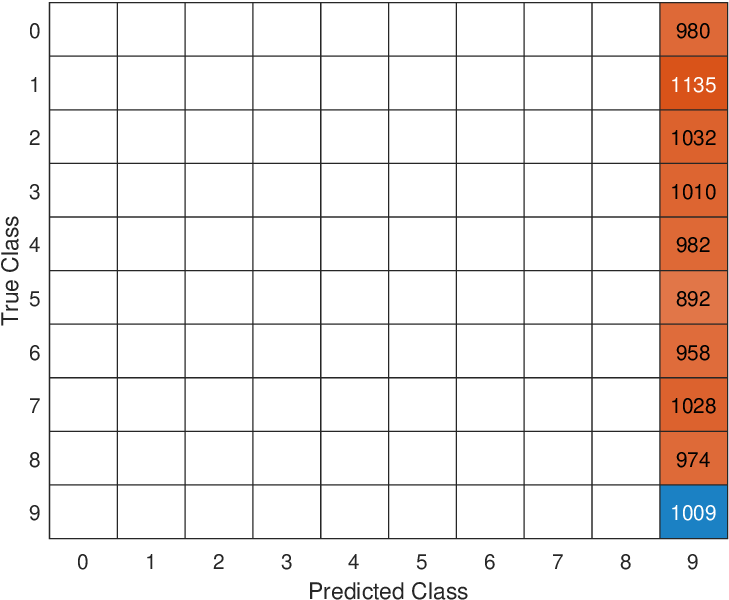} & \includegraphics[width=7.2cm,height=6cm]{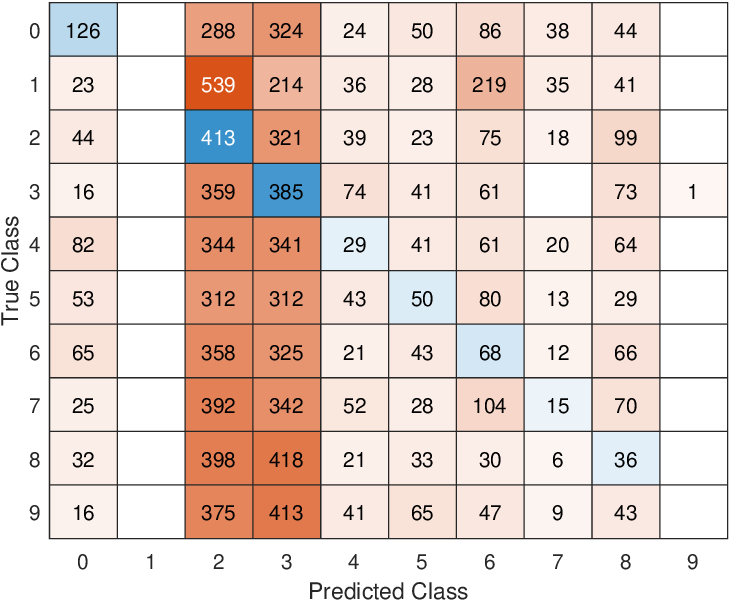}
	\end{tabular}
	\caption{confusion charts for different methods with a compression rate of $20 \%$}
	\label{fig:NN_20_comparison}
\end{figure}

To gain further insight, we examine the confusion matrix of each method for a compression rate of $20 \%$. This is depicted in Figure \ref{fig:NN_20_comparison}.


The supercompress is the only method that predicts every number (with a certain degree of accuracy). The robust version fails to predict zeros. Consequently, it overpredicts other numbers, particularly high numbers such as $7$ and $8$. This is the primary reason for the lower accuracy compared to the normal supercompress. The $K$-means clustering method drastically overestimates $2$ and $3$ to be the correct numbers. This systematic error highlights the ineffectiveness of this method in this context. The most unsatisfactory result is produced by QMC-averaging, which predicts only one number. This is not surprising, given that only two points have notable weights and therefore dominate the loss function. The reason for the absence of certain classes in the prediction can be explained by Table \ref{tab:6}, which shows the distribution of the corresponding responses among the original and compressed data sets. It should be noted that due to rounding, the percentages may not sum to $100$. Additionally, there is a greater concentration of mass around high numbers. This is a consequence of the fact that many clusters contain two points. If the average value has a $0.5$ part, the response is rounded up, for example, the response corresponding to the set $\{8,9\}$ is 9.

\begin{table}[h!]
	\begin{center}
		\begin{tabular}{>{\centering}m{1.9cm}|m{2cm}|m{2cm}|m{2cm}|m{2cm}|m{2cm}|m{2cm}|m{2cm}|m{2cm}|m{2cm}|m{2cm}|}
			\multicolumn{1}{c|}{method} & \multicolumn{1}{c}{0} & \multicolumn{1}{c}{1} & \multicolumn{1}{c}{2}  & \multicolumn{1}{c}{3} & \multicolumn{1}{c}{4} & \multicolumn{1}{c}{5}  & \multicolumn{1}{c}{6}  & \multicolumn{1}{c}{7} & \multicolumn{1}{c}{8} & \multicolumn{1}{c}{9}\\
			\cmidrule(r){1-11}
			\multicolumn{1}{c|}{\multirow{1}{*}{no compression}} & \multicolumn{1}{c}{$10 \%$} & \multicolumn{1}{c}{$11 \%$} & \multicolumn{1}{c}{$10 \%$}  & \multicolumn{1}{c}{$10 \%$} & \multicolumn{1}{c}{$10 \%$} & \multicolumn{1}{c}{$9 \%$}  & \multicolumn{1}{c}{$10 \%$} & \multicolumn{1}{c}{$10 \%$} & \multicolumn{1}{c}{$9 \%$} & \multicolumn{1}{c}{$10 \%$} \\
			\cmidrule(r){1-11}
			\multicolumn{1}{c|}{\multirow{1}{*}{supercompress}} & \multicolumn{1}{c}{$7 \%$} & \multicolumn{1}{c}{$6 \%$} & \multicolumn{1}{c}{$8 \%$}  & \multicolumn{1}{c}{$11 \%$} & \multicolumn{1}{c}{$13 \%$} & \multicolumn{1}{c}{$8 \%$}  & \multicolumn{1}{c}{$7 \%$} & \multicolumn{1}{c}{$9 \%$} & \multicolumn{1}{c}{$15 \%$} & \multicolumn{1}{c}{$15 \%$} \\
			\cmidrule(r){1-11}
			\multicolumn{1}{c|}{\multirow{1}{*}{robust supercompress}} & \multicolumn{1}{c}{$1 \%$} & \multicolumn{1}{c}{$2 \%$} & \multicolumn{1}{c}{$3 \%$}  & \multicolumn{1}{c}{$6 \%$} & \multicolumn{1}{c}{$7 \%$} & \multicolumn{1}{c}{$11 \%$}  & \multicolumn{1}{c}{$15 \%$} & \multicolumn{1}{c}{$18 \%$} & \multicolumn{1}{c}{$19 \%$} & \multicolumn{1}{c}{$20 \%$} \\
			\cmidrule(r){1-11}						
			\multicolumn{1}{c|}{\multirow{1}{*}{$K$-means}} & \multicolumn{1}{c}{$10 \%$} & \multicolumn{1}{c}{$4 \%$} & \multicolumn{1}{c}{$14 \%$}  & \multicolumn{1}{c}{$11 \%$} & \multicolumn{1}{c}{$11 \%$} & \multicolumn{1}{c}{$11 \%$}  & \multicolumn{1}{c}{$12 \%$} & \multicolumn{1}{c}{$10 \%$} & \multicolumn{1}{c}{$12 \%$} & \multicolumn{1}{c}{$6 \%$} \\
			\cmidrule(r){1-11}
		\end{tabular}
		\caption{distribution of the handwritten numbers for a compression rate of $20 \%$}
		\label{tab:6}
	\end{center}
\end{table}

Clearly, a neural network trained with a non-representative data set will be not useful for prediction.


\smallskip
\smallskip

\begin{figure}[h]
	\centering
	\begin{tabular}{c c}
		\textbf{\ \ \ \ 5\%} & \textbf{\ \ \ \ 10\%} \\
		\includegraphics[width=7.2cm,height=6cm]{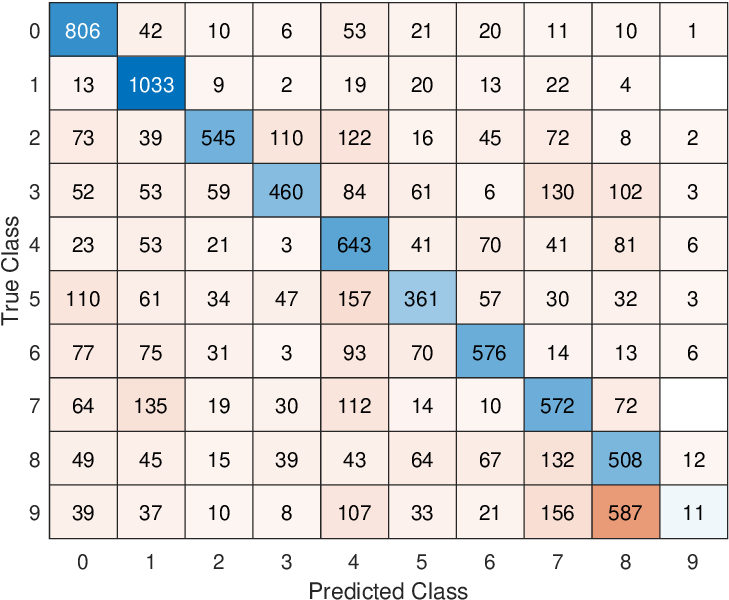} & \includegraphics[width=7.2cm,height=6cm]{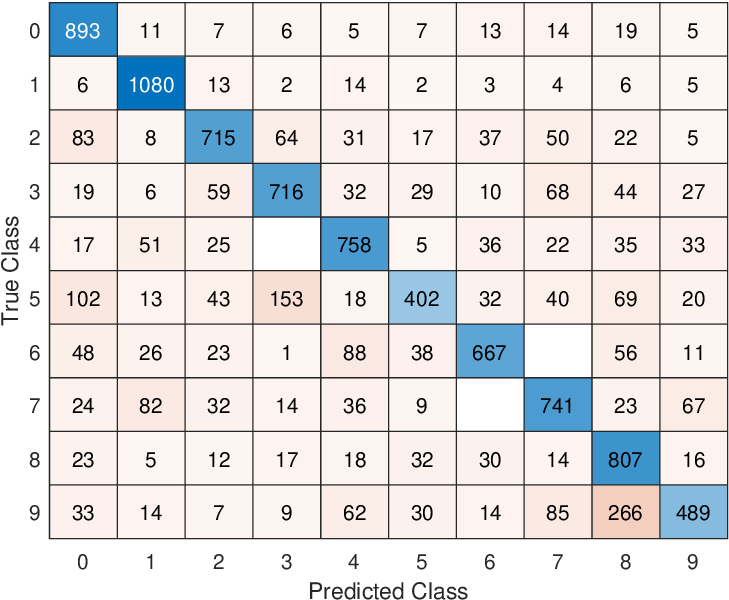}\\
		\textbf{\ \ \ \ 20\%} & \textbf{\ \ \ \ 40\%} \\
		\includegraphics[width=7.2cm,height=6cm]{results/NN_confusionchart_supercom_20.eps} & \includegraphics[width=7.2cm,height=6cm]{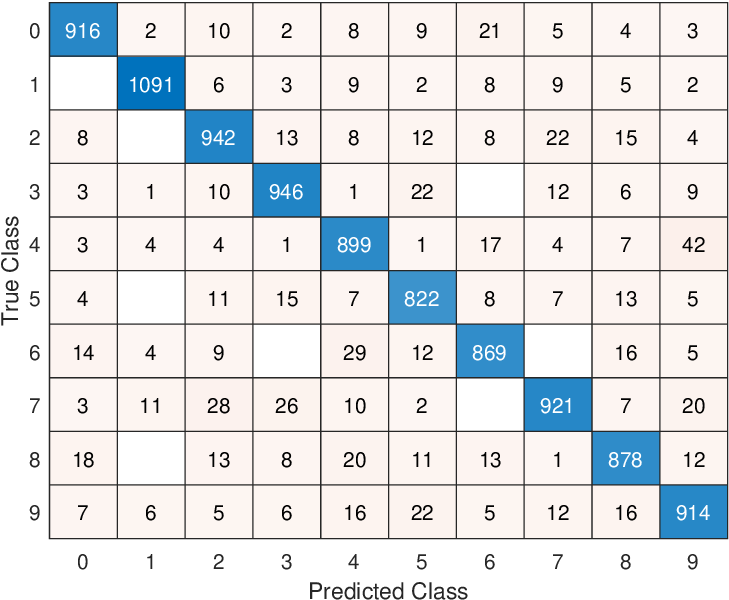}
	\end{tabular}
	\caption{confusion charts for supercompress with different rates}
	\label{fig:NN_supercom_comparison}
\end{figure}

At the end of this section, we present a comparison of the results obtained for different compression rates for the most effective method, namely supercompress. Figure \ref{fig:NN_supercom_comparison} shows the confusion charts. 


The supercompress neural network encounters difficulties in predicting $9$ when the compression rate is low. Instead, it assumes that the value is $8$. However, this misclassification disappears as the number of compressed data points increases. In fact, the result for a compression rate of $20\%$ is already close to the uncompressed prediction. The time required is reduced by approximately $80\%$, while the accuracy only decreases by $4\%$. Perhaps even more impressive is the result for $40\%$. Here, we require approximately half the time, yet the accuracy differs by less than $0.2\%$. This suggests that the supercompress method is a suitable alternative for training the neural network with a smaller set, while maintaining a high level of prediction accuracy.

\section{Conclusion}
\label{sec:4}
In this experimental study we compared two QMC-based data reduction approaches with the supercompress methods from \cite{supercompress}. While supercompress performs well for the classical  MNIST data set, both QMC-methods drastically fail for this data set. A reduced data set given by QMC-points is not able to recapture the original data adequately, neither in combination with an averaging procedure and appropriate weights as proposed in \cite{1} nor in combination with clustering  in the $\boldsymbol{x}$-space as in the QMC-Voronoi method. For simpler test functions, i.e.~more regular problems, the QMC-Voronoi method performs comparably well to the supercompress method, while the QMC-averaging method also performs badly here, in view of both error and running time. The high running time in comparison to the other methods is a direct consequence of the averaging procedure, and this procedure might also explain the bad error behavior: alternating averaging procedures, which even involve binomial coefficients, are prone to  possible numerical instabilities; see e.g. \cite{IEEE,Higham}.

The supercompress method performs best both for regular and irregular data. 
The reason for this seems to be its particular focus on the $y$-space for the clustering; in contrast to it robust version, which uses also the $\boldsymbol{x}$-space, or the QMC-Voronoi method, which only uses the  $\boldsymbol{x}$-space.

Although a general judgement on the applicability of QMC points in date reduction might be premature,  our observations underline the importance of two pieces of wisdom from applied mathematics' folklore:
Regular problems can be solved with non-adaptive methods, while irregular problems require adaptive methods. Moreover, in practical applications, heuristic algorithms can be superior to theoretically well-understood methods.

\newpage

\section{Declarations}

\subsection*{Acknowledgments}%
The publication of this article was funded by the Ministry of Science, Research and the Arts
Baden-W\"urttemberg and the University of Mannheim.

\subsection*{Funding}
This work was financially supported by the DFG Projects GO1920/11-1 and 12-1.

\subsection*{Abbreviations}
QMC, Quasi-Monte Carlo; NP, complexity class; MNIST, Modified National Institute of Standards and Technology.

\subsection*{Availability of data and materials}
Relevant data is cited or drawn randomly.

\subsection*{Competing interests}
The authors declare that they have no competing interests.

\subsection*{Authors’ contributions}
AN and JH worked on the theoretical contribution. The numerical simulation is provided by JH while SG 
contributed to the interpretation of the resulting data.
All authors read and approved the final manuscript.

\section{Appendix}

\appendix
\begin{table}[h!]
	\begin{center}
		\newcommand{\resultrow}[7]{#1 & #2 & #3 & #4 & #6 & #5 & #7\\}
		\begin{tabular}{>{\centering}m{1.9cm}|m{0.65cm}|m{0.65cm}|m{0.6cm}|m{2cm}|m{2cm}|m{2cm}|}
			\multicolumn{1}{c|}{compression rate} & \multicolumn{1}{c}{$N$} & \multicolumn{1}{c}{$s$} & \multicolumn{1}{c|}{$K$} & \multicolumn{1}{c}{$f_1$} & \multicolumn{1}{c}{$f_2$} & \multicolumn{1}{c}{$f_3$}\\
			\cmidrule(r){1-7}
			\multicolumn{1}{c|}{\multirow{2}{*}{$1 \%$}} & \multicolumn{1}{c}{\multirow{2}{*}{$3000$}} & \multicolumn{1}{c}{\multirow{2}{*}{$2$}}  & \multicolumn{1}{c|}{\multirow{2}{*}{$2^{5}$}} & \multicolumn{1}{c}{$3,1054\cdot10^{-4}$}  & \multicolumn{1}{c}{$1,0049\cdot 10^{0}$}  & \multicolumn{1}{c}{$4,6801\cdot10^{-2}$}\\ & \multicolumn{1}{c}{} & \multicolumn{1}{c}{} & \multicolumn{1}{c|}{} & \multicolumn{1}{c}{$2,24\cdot10^{-2}$}  & \multicolumn{1}{c}{$2,0299\cdot 10^{0}$}  & \multicolumn{1}{c}{$3,5859\cdot10^{0}$}\\
			\cmidrule(r){1-7}
			\multicolumn{1}{c|}{\multirow{2}{*}{$4 \%$}} & \multicolumn{1}{c}{\multirow{2}{*}{$3000$}} & \multicolumn{1}{c}{\multirow{2}{*}{$2$}}  & \multicolumn{1}{c|}{\multirow{2}{*}{$2^{7}$}} & \multicolumn{1}{c}{$3,3239\cdot10^{-4}$}  & \multicolumn{1}{c}{$5,5982\cdot 10^{-2}$}  & \multicolumn{1}{c}{$3,993\cdot10^{-5}$}\\ & \multicolumn{1}{c}{} & \multicolumn{1}{c}{} & \multicolumn{1}{c|}{} & \multicolumn{1}{c}{$1,37\cdot10^{-2}$}  & \multicolumn{1}{c}{$9,2858\cdot 10^{-1}$}  & \multicolumn{1}{c}{$3,3122\cdot10^{-2}$}\\
			\cmidrule(r){1-7}
			\multicolumn{1}{c|}{\multirow{2}{*}{$9 \%$}} & \multicolumn{1}{c}{\multirow{2}{*}{$3000$}} & \multicolumn{1}{c}{\multirow{2}{*}{$2$}}  & \multicolumn{1}{c|}{\multirow{2}{*}{$2^{8}$}} & \multicolumn{1}{c}{$2,5366\cdot10^{-4}$}  & \multicolumn{1}{c}{$1,3326\cdot 10^{-2}$}  & \multicolumn{1}{c}{$9,4435\cdot10^{-6}$}\\ & \multicolumn{1}{c}{} & \multicolumn{1}{c}{} & \multicolumn{1}{c|}{} & \multicolumn{1}{c}{$1,9\cdot10^{-3}$}  & \multicolumn{1}{c}{$3,1577\cdot 10^{-1}$}  & \multicolumn{1}{c}{$1,8061\cdot10^{-2}$}\\
			\cmidrule(r){1-7}
			\multicolumn{1}{c|}{\multirow{2}{*}{$17 \%$}} & \multicolumn{1}{c}{\multirow{2}{*}{$3000$}} & \multicolumn{1}{c}{\multirow{2}{*}{$2$}}  & \multicolumn{1}{c|}{\multirow{2}{*}{$2^{9}$}} & \multicolumn{1}{c}{$1,7051\cdot10^{-4}$}  & \multicolumn{1}{c}{$3,9779\cdot 10^{-3}$}  & \multicolumn{1}{c}{$2,0192\cdot10^{-6}$}\\ & \multicolumn{1}{c}{} & \multicolumn{1}{c}{} & \multicolumn{1}{c|}{} & \multicolumn{1}{c}{$1,9\cdot10^{-3}$}  & \multicolumn{1}{c}{$2,9304 \cdot 10^{-1}$}  & \multicolumn{1}{c}{$1,848\cdot10^{-2}$}\\
			\cmidrule(r){1-7}
			\multicolumn{1}{c|}{\multirow{2}{*}{$34 \%$}} & \multicolumn{1}{c}{\multirow{2}{*}{$3000$}} & \multicolumn{1}{c}{\multirow{2}{*}{$2$}}  & \multicolumn{1}{c|}{\multirow{2}{*}{$2^{10}$}} & \multicolumn{1}{c}{$1,0473\cdot10^{-4}$}  & \multicolumn{1}{c}{$9,0664 \cdot 10^{-4}$}  & \multicolumn{1}{c}{$2,4926\cdot10^{-7}$}\\ & \multicolumn{1}{c}{} & \multicolumn{1}{c}{} & \multicolumn{1}{c|}{} & \multicolumn{1}{c}{$2\cdot10^{-3}$}  & \multicolumn{1}{c}{$3,7659\cdot 10^{-1}$}  & \multicolumn{1}{c}{$7,7036\cdot10^{-3}$}\\
			\cmidrule(r){1-7}
		\end{tabular}
		\caption{scaled average error (top row supercompress, bottom row QMC-averaging)}
		\label{tab:1_scaled}
	\end{center}
\end{table}

\begin{table}[h!]
	\begin{center}
		\begin{tabular}{>{\centering}m{1.9cm}|m{0.65cm}|m{0.65cm}|m{0.6cm}|m{2cm}|m{2cm}|m{2cm}|}
			\multicolumn{1}{c|}{compression rate} & \multicolumn{1}{c}{$N$} & \multicolumn{1}{c}{$s$} & \multicolumn{1}{c|}{$K$} & \multicolumn{1}{c}{$f_1$} & \multicolumn{1}{c}{$f_2$} & \multicolumn{1}{c}{$f_3$}\\
			\cmidrule(r){1-7}
			\multicolumn{1}{c|}{\multirow{2}{*}{$9 \%$}} & \multicolumn{1}{c}{\multirow{2}{*}{$3000$}} & \multicolumn{1}{c}{\multirow{2}{*}{$2$}}  & \multicolumn{1}{c|}{\multirow{2}{*}{$2^{8}$}} & \multicolumn{1}{c}{$2,5366\cdot10^{-4}$}  & \multicolumn{1}{c}{$1,3326\cdot 10^{-2}$}  & \multicolumn{1}{c}{$9,4435\cdot10^{-6}$}\\ & \multicolumn{1}{c}{} & \multicolumn{1}{c}{} & \multicolumn{1}{c|}{} & \multicolumn{1}{c}{$1,9\cdot10^{-3}$}  & \multicolumn{1}{c}{$3,1577\cdot 10^{-1}$}  & \multicolumn{1}{c}{$1,8061\cdot10^{-2}$}\\
			\cmidrule(r){1-7}
			\multicolumn{1}{c|}{\multirow{2}{*}{$9 \%$}} & \multicolumn{1}{c}{\multirow{2}{*}{$3000$}} & \multicolumn{1}{c}{\multirow{2}{*}{$5$}}  & \multicolumn{1}{c|}{\multirow{2}{*}{$2^{8}$}} & \multicolumn{1}{c}{$3,286\cdot10^{-4}$}  & \multicolumn{1}{c}{$6,0731\cdot 10^{2}$}  & \multicolumn{1}{c}{$5,7063\cdot 10^{-5}$}\\ & \multicolumn{1}{c}{} & \multicolumn{1}{c}{} & \multicolumn{1}{c|}{} & \multicolumn{1}{c}{$2,3514\cdot10^{-4}$}  & \multicolumn{1}{c}{$9,4799\cdot 10^{3}$}  & \multicolumn{1}{c}{$3,5184\cdot 10^{-4}$}\\
			\cmidrule(r){1-7}
			\multicolumn{1}{c|}{\multirow{2}{*}{$9 \%$}} & \multicolumn{1}{c}{\multirow{2}{*}{$3000$}} & \multicolumn{1}{c}{\multirow{2}{*}{$10$}}  & \multicolumn{1}{c|}{\multirow{2}{*}{$2^{8}$}} & \multicolumn{1}{c}{$3,5688\cdot10^{-4}$}  & \multicolumn{1}{c}{$1,975\cdot 10^{7}$}  & \multicolumn{1}{c}{$5,822\cdot 10^{-1}$}\\ & \multicolumn{1}{c}{} & \multicolumn{1}{c}{} & \multicolumn{1}{c|}{} & \multicolumn{1}{c}{$5,0307\cdot10^{-8}$}  & \multicolumn{1}{c}{$3,1688 \cdot 10^{20}$}  & \multicolumn{1}{c}{$2,8581 \cdot 10^{-8}$}\\
			\cmidrule(r){1-7}
		\end{tabular}
		\caption{scaled average error (top row supercompress, bottom row QMC-averaging)}
		\label{tab:3_scaled}
	\end{center}
\end{table}

\begin{table}[h!]
	\begin{center}
		\begin{tabular}{>{\centering}m{1.9cm}|m{0.65cm}|m{0.65cm}|m{0.6cm}|m{2cm}|m{2cm}|m{2cm}|}
			\multicolumn{1}{c|}{compression rate} & \multicolumn{1}{c}{$N$} & \multicolumn{1}{c}{$s$} & \multicolumn{1}{c|}{$K$} & \multicolumn{1}{c}{$f_1$} & \multicolumn{1}{c}{$f_2$} & \multicolumn{1}{c}{$f_3$}\\
			\cmidrule(r){1-7}
			\multicolumn{1}{c|}{\multirow{2}{*}{$4 \%$}} & \multicolumn{1}{c}{\multirow{2}{*}{$3000$}} & \multicolumn{1}{c}{\multirow{2}{*}{$2$}}  & \multicolumn{1}{c|}{\multirow{2}{*}{$2^{7}$}} & \multicolumn{1}{c}{$1,37\cdot10^{-2}$}  & \multicolumn{1}{c}{$9,2858\cdot 10^{-1}$}  & \multicolumn{1}{c}{$3,3122\cdot10^{-2}$}\\ & \multicolumn{1}{c}{} & \multicolumn{1}{c}{} & \multicolumn{1}{c|}{} & \multicolumn{1}{c}{$4,4655\cdot10^{-4}$}  & \multicolumn{1}{c}{$1,1997\cdot 10^{-1}$}  & \multicolumn{1}{c}{$1,0932\cdot10^{-3}$}\\
			\cmidrule(r){1-7}
			\multicolumn{1}{c|}{\multirow{2}{*}{$9 \%$}} & \multicolumn{1}{c}{\multirow{2}{*}{$3000$}} & \multicolumn{1}{c}{\multirow{2}{*}{$2$}}  & \multicolumn{1}{c|}{\multirow{2}{*}{$2^{8}$}} & \multicolumn{1}{c}{$1,9\cdot10^{-3}$}  & \multicolumn{1}{c}{$3,1577\cdot 10^{-1}$}  & \multicolumn{1}{c}{$1,8061\cdot10^{-2}$}\\ & \multicolumn{1}{c}{} & \multicolumn{1}{c}{} & \multicolumn{1}{c|}{} & \multicolumn{1}{c}{$7,0245\cdot10^{-5}$}  & \multicolumn{1}{c}{$2,1219\cdot 10^{-2}$}  & \multicolumn{1}{c}{$5,299\cdot10^{-4}$}\\
			\cmidrule(r){1-7}
			\multicolumn{1}{c|}{\multirow{2}{*}{$17 \%$}} & \multicolumn{1}{c}{\multirow{2}{*}{$3000$}} & \multicolumn{1}{c}{\multirow{2}{*}{$2$}}  & \multicolumn{1}{c|}{\multirow{2}{*}{$2^{9}$}} & \multicolumn{1}{c}{$1,9\cdot10^{-3}$}  & \multicolumn{1}{c}{$2,9304 \cdot 10^{-1}$}  & \multicolumn{1}{c}{$1,848\cdot10^{-2}$}\\ & \multicolumn{1}{c}{} & \multicolumn{1}{c}{} & \multicolumn{1}{c|}{} & \multicolumn{1}{c}{$1,7547\cdot10^{-4}$}  & \multicolumn{1}{c}{$6,1046\cdot 10^{-2}$}  & \multicolumn{1}{c}{$3,7982\cdot10^{-4}$}\\
			\cmidrule(r){1-7}
		\end{tabular}
		\caption{scaled average error (top row QMC-averaging, bottom row QMC-Voronoi)}
		\label{tab:4_scaled}
	\end{center}
\end{table}

\end{document}